\documentclass[letterpaper]{article} 
\usepackage{aaai23}  
\usepackage{times}  
\usepackage{helvet}  
\usepackage{courier}  
\usepackage[hyphens]{url}  
\usepackage{graphicx} 
\usepackage{natbib}  
\usepackage{caption} 
\usepackage{algorithm}
\usepackage{algorithmic}
\usepackage{multirow}
\usepackage{booktabs}
\usepackage{subcaption}
\usepackage{color}
\usepackage{amssymb}
\usepackage{mathrsfs}
\usepackage{amsmath}
\usepackage[title]{appendix}

\usepackage{newfloat}
\usepackage{listings}
\DeclareCaptionStyle{ruled}{labelfont=normalfont,labelsep=colon,strut=off} 
\floatstyle{ruled}
\newfloat{listing}{tb}{lst}{}
\floatname{listing}{Listing}
\title{3D-TOGO: Towards Text-Guided Cross-Category
3D Object Generation}
\author {
    Zutao Jiang\textsuperscript{\rm 1 \rm 6} \equalcontrib,
    Guansong Lu \textsuperscript{\rm 2} \equalcontrib,
    Xiaodan Liang \textsuperscript{\rm 3 \rm 4 },\\
    Jihua Zhu \textsuperscript{\rm 1} \thanks{Corresponding authors.},
    Wei Zhang \textsuperscript{\rm 2},
    Xiaojun Chang \textsuperscript{\rm 5},
    Hang Xu \textsuperscript{\rm 2 \dag}
}
\affiliations {
    \textsuperscript{\rm 1} School of Software Engineering, Xi’an Jiaotong University
    \textsuperscript{\rm 2} Huawei Noah's  Ark Lab  \\
    \textsuperscript{\rm 3} Sun Yat-sen University 
    \textsuperscript{\rm 4} MBZUAI
    \textsuperscript{\rm 5} ReLER, AAII, University of Technology Sydney
    \textsuperscript{\rm 6} PengCheng Laboratory\\
    taozujiang@gmail.com, luguansong@huawei.com, xdliang328@gmail.com, \\
    zhujh@xjtu.edu.cn, wz.zhang@huawei.com, xiaojun.chang@uts.edu.au, chromexbjxh@gmail.com
}
\usepackage{bibentry}

\begin{document}

\maketitle

\begin{abstract}
Text-guided 3D object generation aims to generate 3D objects described by user-defined captions, which paves a flexible way to visualize what we imagined. Although some works have been devoted to solving this challenging task, these works either utilize some explicit 3D representations (e.g., mesh), which lack texture and require post-processing for rendering photo-realistic views; or require individual time-consuming optimization for every single case. Here, we make the first attempt to achieve generic text-guided cross-category 3D object generation via a new 3D-TOGO model, which integrates a text-to-views generation module and a views-to-3D generation module. The text-to-views generation module is designed to generate different views of the target 3D object given an input caption. $prior$-guidance, caption-guidance and view contrastive learning are proposed for achieving better view-consistency and caption similarity. Meanwhile, a pixelNeRF model is adopted for the views-to-3D generation module to obtain the implicit 3D neural representation from the previously-generated views.
Our 3D-TOGO model generates 3D objects in the form of the neural radiance field with good texture and requires no time-cost optimization for every single caption. Besides, 3D-TOGO can control the category, color and shape of generated 3D objects with the input caption.
Extensive experiments on the largest 3D object dataset (i.e., ABO) are conducted to verify that 3D-TOGO can better generate high-quality 3D objects according to the input captions across \textbf{98} different categories, in terms of PSNR, SSIM, LPIPS and CLIP-score, compared with text-NeRF and Dreamfields.
\end{abstract}

\section{Introduction}
\label{sec:introduction}

\looseness=-1
Automatic 3D object generation has significant application values for many practical application scenarios, including games, movies, virtual reality, etc. In this paper, we study a challenging yet interesting and valuable task, called text-guided 3D object generation. With a text-guided 3D object generation model, one can give a textual description of their wanted 3D object, and leverage such a model to generate the corresponding 3D object, providing a flexible path for visualizing what we imagined.

\begin{figure}[t]
\centering
\includegraphics[width=0.85\linewidth]{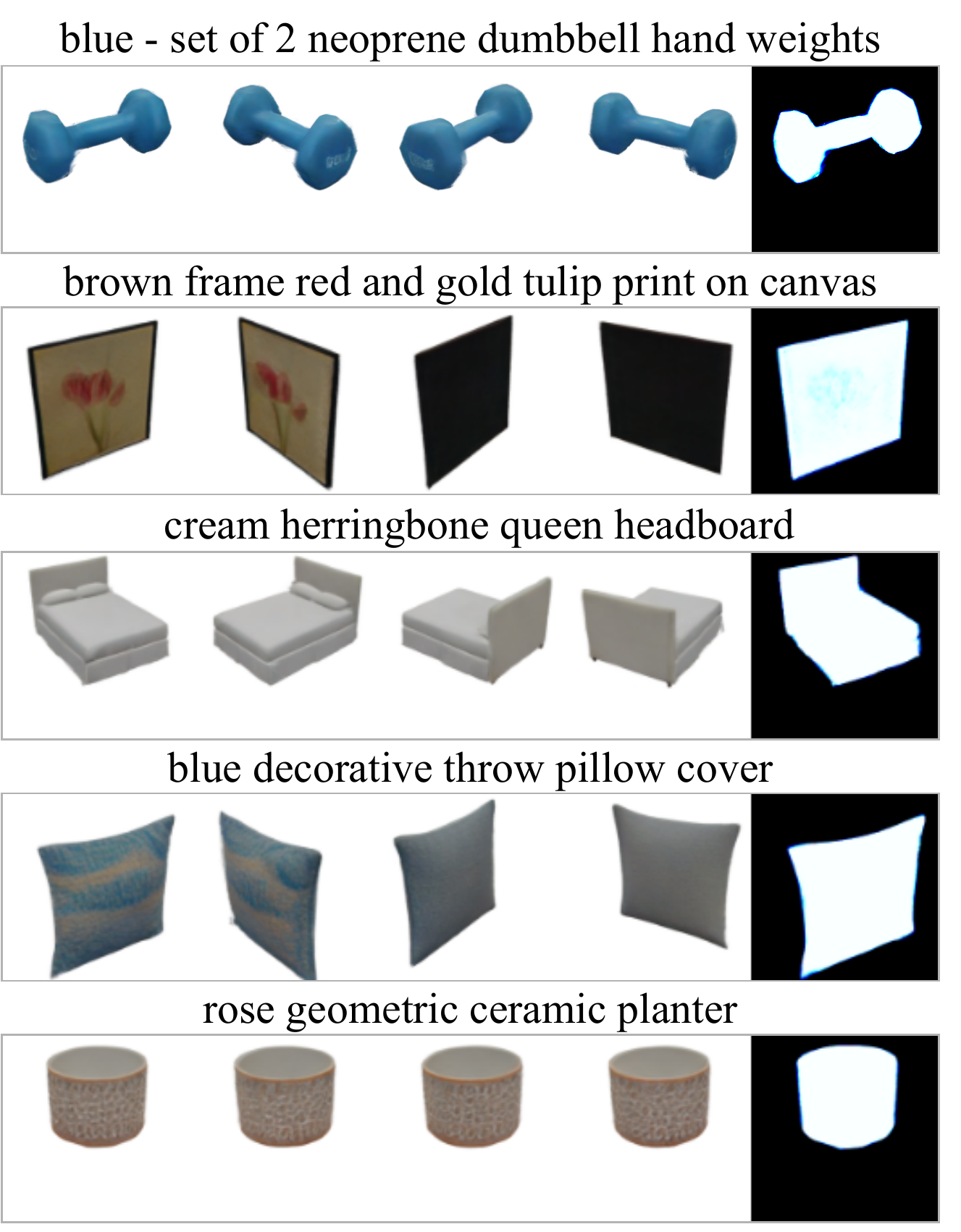}
\vspace{-2mm}
\caption{Generation results of our proposed 3D-TOGO model. For each case, 
we show the input caption, 4 rendered novel views of the generated 3D object and the transmittance from the first view. 
Transmittance represents how visible a point is from a particular view.
}
\label{fig:main_results}
\end{figure}

\looseness=-1
Along with the success of image generation models \cite{goodfellow2014gan,vaswani2017attention,ho2020ddpm}, plenty of works have been devoted for text-to-image generation \cite{GAN-INT-CLS,zhang2017stackgan,zhang2018stackgan++,xu2018attngan,sisgan,dalle,ding2021cogview,ding2022cogview2,nichol2021glide,ramesh2022dalle2}, which shows appealing text-guided generation results. However, there are still few works for text-guided 3D object generation. Some prior works generate 3D shapes from natural language descriptions in the form of meshes \cite{text2mesh}, voxels \cite{chen2018text2shape}, point clouds \cite{zhou20213d}, and implicit functions \cite{liu2022implicit-text-3D-generation}. While they provide promising results, the issue is that they require tedious post-processing steps, e.g. unwrapping a UV map in Blender, due to the lack of texture when used for multimedia applications. Recently, Neural Radiance Field (NeRF \cite{mildenhall2020nerf}) has been successfully applied to the novel view synthesis task. Compared with other 3D representations, neural radiance fields can be sampled at high spatial resolutions and is easy to optimize. Empowered with the visual-language alignment capability of the pre-trained CLIP model, Dreamfields \cite{jain2021dreamfields} leverages a given input text to guide the training of neural radiance fields. The shortcoming of Dreamfields is that it requires individually optimizing a network for each input text, which is time-consuming and computation expensive. Built on the disentangled conditional NeRF \cite{schwarz2020graf} and CLIP model, CLIP-NeRF \cite{wang2021clip} designs two code mappers to edit the shape and color of {existing} 3D objects with a text or image prompt. However, it only allows editing objects in the same category.

\looseness=-1
In this paper, we make the first attempt to achieve the generic text-guided \textbf{cross-category} 3D object generation and propose our 3D-TOGO model, standing on the progress of text-to-image generation models and Neural Radiance Fields. Our 3D-TOGO model consists of two modules: a) a view-consistent text-to-views generation module that generates views of the target 3D object given an input caption; b) a generic views-to-3D generation module for 3D object generation based on the previously-generated views. Specifically, we adopt the Transformer-based auto-regressive model \cite{vaswani2017attention} for our text-to-views generation module, because of its excellent cross-modal fusion capability. To complement the original token-level cross-entropy loss, we introduce the fine-grained pixel-level supervision signals for better view fidelity. 
We also incorporate caption-guidance by leveraging the visual-language alignment ability from CLIP \cite{radford2021clip} for better caption similarity. Furthermore, our text-to-views generation module achieves better view-consistency by conditioning on $prior$ view and adopting a novel view contrastive learning method.
For the generic views-to-3D generation module, we follow the diagram of the previously-proposed pixelNeRF \cite{yu2021pixelnerf}. As pixelNeRF optimizes neural radiance fields in the view space of the input image, so we can process each generated view independently and obtain an individual latent intermediate representation for each generated view, which can be aggregated across different views and generate the desired 3D neural representation matched with the input text. Besides, our views-to-3D generation module aims to learn scene prior instead of remembering the training dataset, allowing it to be used for generating objects across different categories.

\looseness=-1
We perform extensive experiments on the largest 3D object dataset ABO \cite{collins2021abo}. Quantitative and qualitative comparisons against baseline methods, including text-NeRF and Dreamfields \cite{jain2021dreamfields}, show that our proposed 3D-TOGO model can better generate high-quality 3D objects according to the input captions across different object categories. Compared to baseline methods, the average CLIP-score of our model surpasses \textbf{4.4} on randomly selected text inputs, 
indicating better semantic consistency between the input captions and the generated 3D objects of our model.
Besides, results from our 3D-TOGO model show that text-guided 3D object generation allows for flexible control over categories, colors and shapes. Our main contributions are summarized as follows:

\begin{itemize}
\item We make the first attempt to resolve the new text-guided cross-category 3D object generation problem and propose 3D-TOGO model, which has an efficient generation process requiring no inference time optimization.

\item We propose a text-to-views generation module to generate consistent views given the input captions. 
We design $prior$-guidance to improve the consistency between adjacent views of a 3D object and introduce view contrastive learning to improve the consistency between different views of a 3D object. Caption-guidance is proposed for better caption similarity. Fine-grained pixel-level supervision is designed for better view fidelity.

\item Our 3D-TOGO model can generate high-quality 3D objects across \textbf{98} categories. Besides, our 3D-TOGO model is empowered with the ability to control the \textbf{category}, \textbf{color} and \textbf{shape} according to the input caption.
\end{itemize}

\section{Related Work}
\label{sec:related_work}
\textbf{Text-to-Image Generation.} Text-to-image generation focuses on generating images described by input captions. Based on the progresses on generative models, including generative adversarial networks (GANs \cite{goodfellow2014gan}), auto-regressive model \cite{vaswani2017attention} and diffusion model \cite{ho2020ddpm}, there are numbers of works for text-to-image generation. Among them, 
many GAN-based models are proposed for better visual fidelity and caption similarity \cite{GAN-INT-CLS,zhang2017stackgan,zhang2018stackgan++,xu2018attngan,li2019controllable,sisgan,tao2020dfgan,ye2021xmcgan}. 
However, GANs suffer from the well-known problem of mode-collapse and unstable training process. 
Besides GANs, another line of works explore applying Transformer-based auto-regressive model for text-to-image generation \cite{dalle,ding2021cogview,esser2021imagebart,ding2022cogview2,zhang2021m6-ufc,lee2022rq-vae}. 
Recent works adopt diffusion model for text-to-image generation \cite{nichol2021glide,ramesh2022dalle2}. 
However, as the diffusion model predicts the added noise instead of the target images, it is complicated to apply constraints on the generated images.
We adopt the architecture of Transformer \cite{vaswani2017attention} for our view-consistent text-to-views generation module, due to its high cross-modality fusion capability proven in the domain of multi-modal pre-training \cite{li2019visualbert,chen2020uniter,wang2021simvlm,tan2019lxmert,ni2021m3p,zhou2021uc2,zhuge2021kaleido} and generation mentioned above. Furthermore, we adopt the auto-regressive generation paradigm due to the aforementioned drawbacks of GANs and diffusion model.

\textbf{Text-Guided 3D Object Generation.} 
\looseness=-1
Compared with text-to-image generation, it is more challenging to generate 3D objects from the given text description. Some early works generate or edit 3D objects with a pre-trained CLIP model \cite{radford2021clip}. 
Text2Mesh\cite{text2mesh} edits the style of a 3D mesh by conforming the rendered images to a target text prompt with a CLIP-based semantic loss. It is tailored to a single mesh and could not generate 3D objects from scratch given a text prompt.
\cite{khalid2022text} facilitates zero-shot text-driven mesh generation by deforming from a template mesh guided by CLIP.
Text2shape \cite{chen2018text2shape} generates the voxelized objects using text-conditional Wasserstein GAN \cite{arjovsky2017wasserstein}, but only allows the 3D object generation of individual category and the performance is limited by the low-resolution 3D representation. CLIP-Forge \cite{sanghi2021clip-forge} models the distribution of shape embeddings conditioned on the image features using a normalizing flow network during the training stage, and then conditions the normalizing flow network with text features to generate a shape embedding during the inference stage, which can be converted into a 3D shape via the shape decoder. However, their generated 3D objects lack color and texture and the quality of generated 3D objects is still limited, which is crucial for practical applications. 
Recently, \cite{liu2022implicit-text-3D-generation} 
represents 3D shape with the implicit occupancy representation, which can be used to predict an occupancy field. They design a cyclic loss to encourage the consistency between the generated 3D shape and the input text. However, it cannot generate realistic 3D objects with high fidelity. There are also some works focus on 3D avatar generation and animation from text\cite{hong2022avatarclip,hu2021text,canfes2022text}, while our work focuses on 3D object generation from text. Compared with the above approaches, our 3D-TOGO model can generate high-quality 3D objects with color and texture across categories and requires no inference time optimization.

\section{Method}
\label{sec:method}

\begin{figure*}[t]
\centering
\includegraphics[width=0.9\linewidth]{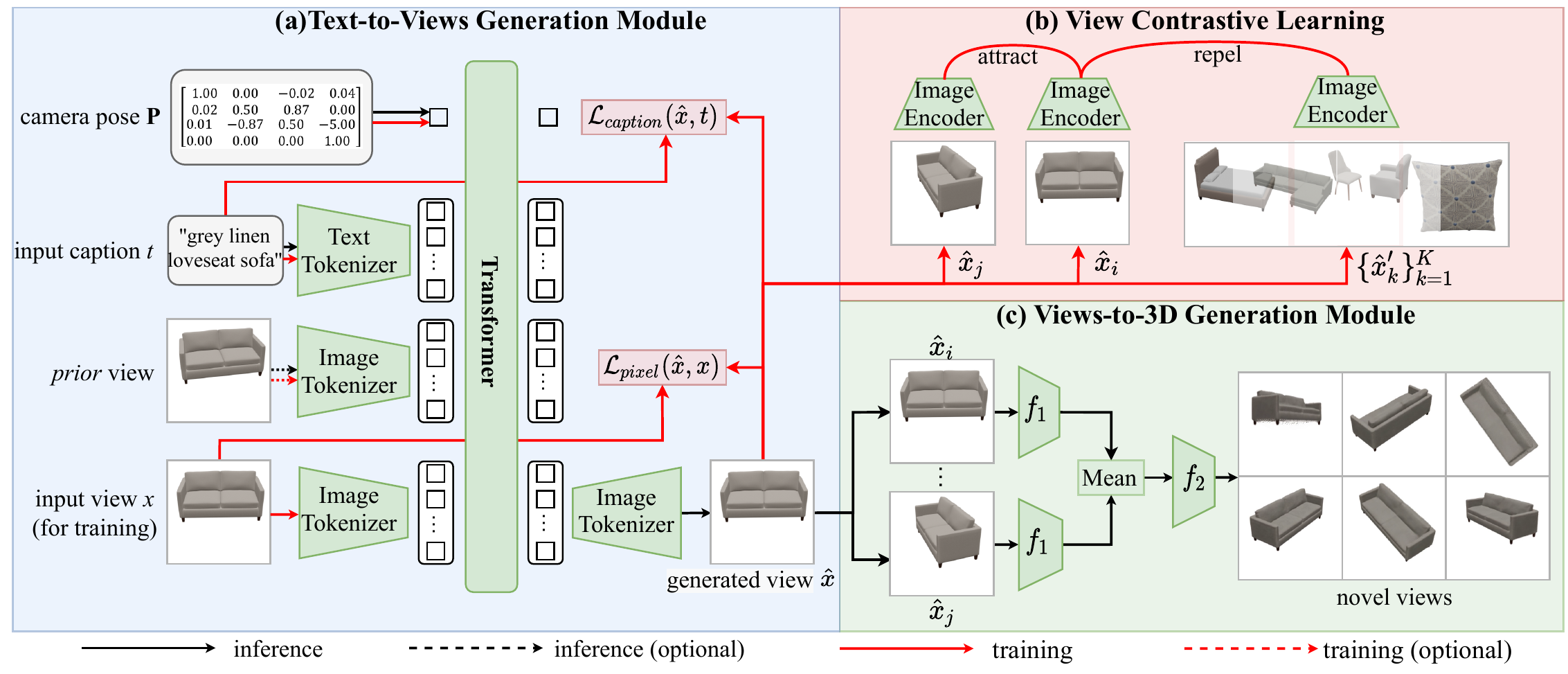}
\caption{The framework of our 3D-TOGO model for text-guided 3D object generation. 
(a) Text-to-views generation module.
Fine-grained pixel-level supervision signal $\mathcal{L}_{pixel}$ and caption-guidance loss $\mathcal{L}_{caption}$ are for better view fidelity and caption similarity. (b) View contrastive learning for better view-consistency.
(c) Views-to-3D generation module. It takes the previously-generated views as input and generates the implicit 3D neural representation, from which novel views can be obtained.
}
\label{fig:framework}
\vspace{-4mm}
\end{figure*}

Figure \ref{fig:framework} shows the framework of our proposed 3D-TOGO model for text-guided 3D object generation. In this section, we will first introduce our view-consistent text-to-views generation module, which takes captions of 3D objects and different camera poses as input, enabling the multi-view image generation.
Then we will introduce how to obtain the 3D implicit neural representation of the objects from the previously-generated views.

\subsection{View-Consistent Text-to-Views Generation Module}
\label{sec:View-consistent Text-to-views Generation Model}

Given an input caption $t$, our text-to-views generation module aims to generate 2D images $\{ {\hat x_i}\} _{i = 1}^N$ of different camera poses $ \{ {{\bf{P}}_i}\} _{i = 1}^N$ for the corresponding 3D object described by caption $t$, where $N$ is the number of generated views for a 3D object. $\{ {x_i}\} _{i = 1}^N$ denotes the ground truth images in dataset. The generated images $\hat{x}_i$ need to be consistent with its corresponding input caption $t$ and camera pose ${\bf{P}}_i$. Besides, different views of the same input caption need to be view-consistent, i.e., images of different poses need to be consistent with each other as they are rendered from the same 3D object. For brevity, we omit the subscript of $x$, $\hat{x}$, and $\bf{P}$ in the rest of this section.

\subsubsection{Base Text-to-Views Generation Module}
To generate images of different camera poses given an input caption, we start with designing our base generation module to generate image $\hat x$ conditioned on caption $t$ and camera pose ${\bf{P}}$.
We adopt the architecture of Transformer \cite{vaswani2017attention} due to its high cross-modality fusion capability \cite{li2019visualbert,wang2021simvlm,dalle,ding2021cogview,lee2022rq-vae}. 
Following previous works, we transform camera pose ${\bf{P}}$, caption $t$ and image $x$ into sequences of tokens, and train a decoder-only Transformer model with a causal attention mask to predict the sequence of image tokens autoregressively conditioned on camera pose and caption tokens. 

Specifically, our Transformer-based image generation module consists of an VQGAN \cite{vqgan} model, serving as an image tokenizer for quantizing the input image as discrete tokens and recovering the origin image from these discrete tokens, and a Transformer model for fitting the joint distribution of camera pose, caption and image tokens. 
The autoencoder model consists of an encoder $E$, a decoder $G$ and a codebook $Z \in \mathbb{R}^{K \times n_z}$ containing $K$ $n_z$-dimensional codes. 
Given an image $x \in \mathbb{R}^{H \times W \times 3}$, $E$ first encodes the image into a 
two-dimensional feature map $\mathbf{F} \in \mathbb{R}^{h \times w \times n_z}$, and then the feature map $\mathbf{F}$ is quantized by replacing each pixel embedding  with its closest code within the codebook element-wisely: $\mathbf{\hat{F}}_{ij} = \arg\min_{\mathbf{z}_k}\parallel \mathbf{F}_{ij} - \mathbf{z}_{k} \parallel^{2}$.
The decoder $G$ is for taking the quantized feature map $\mathbf{\hat{F}}$ as input and reconstructing an pixel-level image $\hat{x}$ close to the original image $x$, 
i.e., $\hat{x} \approx x$. 
With the aforementioned image tokenizer, image $x$ can be tokenized as a sequence of discrete tokens $\{\mathbf{I}_i\}_{i=1}^{N_{\mathbf{I}}}$. Meanwhile, caption $t$ is encoded into sequence of discrete tokens $\{\mathbf{T}_i\}_{i=1}^{N_{\mathbf{T}}}$ with a Byte-Pair Encoding (BPE \cite{sennrich2015bpe}) tokenizer. $N_{\mathbf{I}}$ and $N_{\mathbf{T}}$ denotes the length of image token sequence and caption token sequence respectively. For camera pose $\bf{P}$, we select $N_{\bf{P}}$ poses across different objects so that each camera pose is correspond to an unique token denoted as $\mathbf{V}$. The Transformer is trained to predict the sequence of $[\mathbf{V}, \{\mathbf{T}_i\}_{i=1}^{N_{\mathbf{T}}}, \{\mathbf{I}_i\}_{i=1}^{N_{\mathbf{I}}}]$ auto-regressively, which minimizes cross entropy losses applied to the predicted tokens of camera pose, text and image, respectively as follows:
    $\mathcal{L}_{pose} = CE(\hat{\mathbf{V}}, \mathbf{V}), ~
    \mathcal{L}_{txt} = \mathbb{E}_i[CE(\hat{\mathbf{T}}_i, \mathbf{T}_i)], ~
    \mathcal{L}_{img} = \mathbb{E}_i[CE(\hat{\mathbf{I}}_i, \mathbf{I}_i)],$
where $\hat{\mathbf{V}}$, $\hat{\mathbf{T}_i}$ and $\hat{\mathbf{I}_i}$ are the predicted tokens of camera pose, caption and image respectively; $\mathbb{E}_i[\cdot]$ denotes the expectation and $CE$ represents cross-entropy loss.

\subsubsection{Pixel-Level Supervision and Caption-Guidance}
One shortage of the aforementioned base text-to-views generation module is that the training loss is applied on image tokens, lacking fine-grained pixel-level supervision signals and leading to low visual quality. To this end, to complement such token-level loss, we explore some pixel-level supervision signals applied on the generated image $\hat{x}$ (decoded from the generated image tokens $\{\mathbf{\hat{I}}_i\}_{i=1}^{N_{\mathbf{I}}}$ with the image tokenizer) for more fine-grained supervision signals and better image fidelity. We first explore training losses between the original image ${x}$ and generated image ${\hat{x}}$. In our preliminary experiments, we tried $L1$ loss and Perceptual loss \cite{johnson2016perceptual}. Similar results are observed so that we use the simpler $L1$ loss as: $\mathcal{L}_{pixel} = L1(\hat{x}, x).$
Gradient back-propagation from the generated image ${\hat{x}}$ to the generated image tokens $\{\mathbf{\hat{I}}_i\}_{i=1}^{N_{\mathbf{I}}}$ is implemented with straight-through estimator \cite{bengio2013estimating}.

Besides, we explore some supervision signals applied between the generated image $\hat{x}$ and input caption $t$, called caption-guidance, for better caption similarity, i.e., the generated images better match the semantics of the input captions. To this end, we explore leveraging the power of the CLIP
model \cite{radford2021clip}, which is pre-trained with 400 million image-text pairs collected from the Web and shows excellent zero-shot visual-language alignment capability. Specifically, we utilize the pre-trained ViT-B/32 CLIP model to calculate the similarity score between the generated image ${\hat{x}}$ and input caption $t$ and apply a caption-guidance loss as: $\mathcal{L}_{caption} = - Sim_{CLIP}({\hat{x}}, t),$
to enforce the generation module to generate images that are more semantically similar to the input caption.

\subsubsection{$prior$-Guidance and View Contrastive Learning}
Until now, it is still challenging for the text-to-views generation module to generate view-consistent images among different camera poses, as it is (1) trained to generate images conditioned on only camera pose and caption, without information from images of other poses, (2) without any view-consistency supervision signal during training.

To improve the consistency between adjacent views, we first propose to condition the generation module on a piece of extra information: an image of another camera pose, which we call $prior$ view. To this end, we specify a fixed order of different camera poses and condition image generation of the current camera pose on the previous one. During training, such $prior$ view is masked half of the time so that the generation module is able to perform image generation with and without $prior$ view. During inference, given an input caption $t$, the first view is generated without $prior$ view, while the others are generated using the previously generated one as $prior$ view one by one in order. Tokens of input $prior$ view and predicted $prior$ view are denoted as $\{\mathbf{I}_{i}^{prior}\}_{i=1}^{N_{\mathbf{I}}}$ and $\{\mathbf{\hat{I}}_i^{prior}\}_{i=1}^{N_{\mathbf{I}}}$ respectively. A reconstruction loss is also applied on the predicted $prior$ view tokens as: $\mathcal{L}_{prior} = \mathbb{E}_i ~ CE(\hat{\mathbf{I}}_i^{prior}, \mathbf{I}_i^{prior}).$

Furthermore, we incorporate the concept of contrastive learning for better view-consistency. 
In our case, as we can see, different views of the corresponding object of an input caption should be closer to each other, than to views of the corresponding object of a different input caption. This is the same as the objective of contrastive learning. To this end, we propose view contrastive learning, where views of the same object are treated as positive samples of each other, while views of different objects are treated as negative samples of each other. During training, we generate two different views $\hat{x}_i$, $\hat{x}_j$, $i \neq j$ of the same object, and a set of $K$ views $X = \{\hat{x}_1', \hat{x}_2', ..., \hat{x}_K'\}$ of different objects. Besides, we learn an image encoder $f_{enc}$ for extracting view representations $f_{enc}(x)$. Then the objective function of view contrastive learning can be formulated as follows:
\begin{equation}
    \mathcal{L}_{contrastive} = -\log \frac{\exp(sim(f_{enc}(\hat{x}_i), f_{enc}(\hat{x}_j)) / \tau)}{\sum_{x \in X} \exp(sim(f_{enc}(\hat{x}_i), f_{enc}(x)) / \tau)},
\end{equation}
where $sim(\cdot, \cdot)$ denotes cosine similarity and $\tau$ denotes a temperature parameter.

Finally, the overall objective function of our view-consistent text-to-views generation module is as follows:
\begin{equation}
    \begin{aligned}
    \mathcal{L} = &\lambda_{pose}\mathcal{L}_{pose} + \lambda_{txt}\mathcal{L}_{txt} \\
    &+ \lambda_{prior}\mathcal{L}_{prior} + \lambda_{img}\mathcal{L}_{img}  + \lambda_{pixel}\mathcal{L}_{pixel} \\
    &+ \lambda_{caption}\mathcal{L}_{caption} + \lambda_{contrastive}\mathcal{L}_{contrastive},
    \end{aligned}
\end{equation}
where $\lambda_{pose}$, $\lambda_{txt}$, $\lambda_{prior}$, $\lambda_{img}$, $\lambda_{pixel}$, $\lambda_{caption}$ and $\lambda_{contrastive}$ are the balancing coefficients.

\subsection{Views-to-3D Generation Module}

Given images $\hat{\mathscr{I}} = \left\{ {{{\hat x}_i}} \right\}_{i = 1}^N$ generated by the text-to-views module, the aim of views-to-3D module is to obtain the implicit neural representation of the generated 3D object, where $N$ is the number of generated images. In experiments, we find that NeRF \cite{mildenhall2020nerf} fails to obtain high-quality novel view synthesis results in some cases, if we naively optimize NeRF with the generated images. This is because that there are still some small inconsistent contents among the generated images. Therefore, we introduce pixelNeRF \cite{yu2021pixelnerf} to firstly learn scene prior from the ground-truth images $ \mathscr{I} = \left\{ {{{ x}_i}} \right\}_{i = 1}^M$ across objects in the training data, where $M (M \geq N)$ is the number of rendered images of 3D objects. Please refer to \cite{yu2021pixelnerf} regarding the network architecture details of pixelNeRF model.

Once obtaining the scene prior, we can encode the generated 3D object as a continuous volumetric radiance field $f$ of color $c$ and density $\sigma$ by using the generated multi-view images $\hat{\mathscr{I}}$. Similar to pixelNeRF, we use the view space of the generated images instead of the canonical space. Specially, for a 3D query point $y$ in the neural radiance fields, we first retrieve the corresponding image features from $\hat{\mathscr{I}}$ by $\{ {w_i}\} _{i = 1}^N = \{ {{\bf{W}}_i}({\pi _i}(y))\} _{i = 1}^N$
, where ${{\bf{W}}_i} = E({\hat x_i})$ is the feature volume extracted from the generated image $\hat x_i$, ${\pi _i}(y)$ denotes the corresponding image coordinate on the image plane of the generated image $\hat x_i$, and ${{\mathop{\rm W}\nolimits} _i}({\pi _i}(y))$ represents the image feature extracted from feature volume ${{\mathop{\rm W}\nolimits} _i}$ for the 3D query point $y$. 

\looseness=-1
Then, we need to obtain the intermediate representation $U = \{ {U_i}\} _{i = 1}^N$ in each view space of the generated images 
$\hat{\mathscr{I}}$ for query point $y$ with view direction $d$ as follows: 
\begin{equation}
   {U_i} = {f_1}(\gamma ({\bf{H}}{(y)_i}),{d_i};{w_i}), 
\end{equation}
where ${\bf{H}}{(y)_i} = {{\bf{P}}_i}y = {{\bf{R}}_i}y + {h_i}$ denotes that transforming the query point $y$ into the coordinate system of the generated image $\hat x_i$, ${{\bf{P}}_i}$ is the world to camera transformation matrix, ${{\bf{R}}_i}$ is the rotation matrix, ${h_i}$ is the translation vector, $\gamma ( \cdot )$ represents a positional encoding on the transformed query point ${\bf{H}}{(y)_i}$ with 6 exponentially increasing frequencies \cite{mildenhall2020nerf}; ${d_i} = {{\bf{R}}_i}d$ denotes transforming the view direction $d$ into the coordinate system of the generate image $\hat x_i$; $w_i$ is the corresponding image feature extracted from the generate image $\hat x_i$; ${f_1}( \cdot )$ represents the layers of the pixelNeRF to process transformed query point, transformed view direction and the corresponding extracted image feature in the view space of the generated images $\hat{\mathscr{I}}$ independently, which has been trained on the ABO training dataset.

After obtaining all the intermediate representation $U = \{ {U_i}\} _{i = 1}^N$, we use the average pooling operator $\eta$ to aggregate them and then pass the layers of the pixelNeRF to process the aggregated representation. This process can be written as:
\begin{equation}
    f(y,d) = (\sigma (y),c(y,d)) = {f_2}(\eta ({U_i}))_{i = 1}^N,
\end{equation}
where ${f_2}$ denotes the layers to process the aggregated representation $\eta ({U_i})_{i = 1}^N$, $\sigma (y)$ is the density of the 3D query point $y$ which is independent of the view direction $d$, $c(y,d)$ represents the color of the 3D query point $y$ in the view direction $d$, $f( \cdot )$ is the final continuous volumetric radiance field which representing the generated 3D object matched with the input caption $t$. For the photo-realistic rendering of the generated 3D object, we use the volume rendering technique proposed in \cite{mildenhall2020nerf}.

\section{Experiments}
\label{sec:experiments}

\begin{table*}[]
\footnotesize
\centering
\caption{
Quantitative comparison against text-NeRF (short for text-to-views generation + NeRF\cite{mildenhall2020nerf})   and Dreamfields \cite{jain2021dreamfields}.}
\begin{tabular}{c|c|ccccccc}
\toprule[1pt]
Metric                              & Method      & text1                             & text2                             & text3                             & text4                             & text5                             & text6                             & 12 Texts Avg.  \\ \midrule
\multirow{2}{*}{PSNR $\uparrow$}               & text-NeRF   & 18.15                             & 19.96                             & 0.79                              & 22.02                             & 18.12                             & 0.48                              & 14.04          \\
                                    & Ours        & \textbf{20.12}                    & \textbf{23.34}                    & \textbf{26.41}                    & \textbf{24.02}                    & \textbf{23.80}                    & \textbf{26.53}                    & \textbf{24.98} \\ \midrule
\multirow{2}{*}{SSIM $\uparrow$}               & text-NeRF   & 0.856                             & 0.898                             & 0.001                             & 0.876                             & 0.863                             & 0.001                             & 0.636          \\
                                    & Ours        & \textbf{0.889}                    & \textbf{0.920}                    & \textbf{0.925}                    & \textbf{0.898}                    & \textbf{0.897}                    & \textbf{0.864}                    & \textbf{0.900} \\ \midrule
\multirow{2}{*}{LPIPS $\downarrow$}              & text-NeRF   & 0.138                             & 0.091                             & 0.503                             & 0.142                             & 0.167                             & 0.475                             & 0.239          \\
                                    & Ours        & \textbf{0.122}                    & \textbf{0.063}                    & \textbf{0.075}                    & \textbf{0.128}                    & \textbf{0.165}                    & \textbf{0.082}                    & \textbf{0.092} \\ \midrule
\multirow{3}{*}{CLIP- Score $\uparrow$}        & Dreamfields & 18.92                             & 21.34                             & 18.13                             & 16.31                             & 18.08                             & 18.73                             & 18.40          \\
                                    & text-NeRF   & 26.65                             & 20.28                             & 13.90                             & 21.12                             & 23.16                             & 11.84                             & 18.38          \\
                                    & Ours        & \textbf{27.08}                    & \textbf{22.67}                    & \textbf{20.98}                    & \textbf{22.74}                    & \textbf{24.13}                    & \textbf{25.08}                    & \textbf{22.84} \\ \midrule
\multirow{3}{*}{Object Fidelity $\downarrow$}    & Dreamfields & \multicolumn{1}{r}{3.00}             & \multicolumn{1}{r}{2.56}          & \multicolumn{1}{r}{2.04}          & \multicolumn{1}{r}{2.94}          & \multicolumn{1}{r}{2.94}          & \multicolumn{1}{r}{1.94}          & 2.63           \\
                                    & text-NeRF   & \multicolumn{1}{r}{1.97}          & \multicolumn{1}{r}{2.32}          & \multicolumn{1}{r}{2.93}          & \multicolumn{1}{r}{2.04}          & \multicolumn{1}{r}{1.97}          & \multicolumn{1}{r}{2.95}          & 2.27           \\
                                    & Ours        & \multicolumn{1}{r}{\textbf{1.03}} & \multicolumn{1}{r}{\textbf{1.12}} & \multicolumn{1}{r}{\textbf{1.03}} & \multicolumn{1}{r}{\textbf{1.02}} & \multicolumn{1}{r}{\textbf{1.09}} & \multicolumn{1}{r}{\textbf{1.12}} & \textbf{1.11}  \\ \midrule
\multirow{3}{*}{Caption Similarity $\downarrow$} & Dreamfields & \multicolumn{1}{r}{2.97}          & \multicolumn{1}{r}{2.66}          & \multicolumn{1}{r}{2.02}          & \multicolumn{1}{r}{2.94}          & \multicolumn{1}{r}{2.87}          & \multicolumn{1}{r}{2.01}          & 2.64           \\
                                    & text-NeRF   & \multicolumn{1}{r}{1.98}          & \multicolumn{1}{r}{2.19}          & \multicolumn{1}{r}{2.95}          & \multicolumn{1}{r}{2.03}          & \multicolumn{1}{r}{1.97}          & \multicolumn{1}{r}{2.95}          & 2.25           \\
                                    & Ours        & \multicolumn{1}{r}{\textbf{1.05}} & \multicolumn{1}{r}{\textbf{1.15}} & \multicolumn{1}{r}{\textbf{1.03}} & \multicolumn{1}{r}{\textbf{1.03}} & \multicolumn{1}{r}{\textbf{1.16}} & \multicolumn{1}{r}{\textbf{1.04}} & \textbf{1.11} 
                                    \\
\bottomrule [1pt]
\end{tabular}
\label{table:compare_baseline}
\end{table*}

\textbf{Dataset.} Our approach is evaluated on Amazon-Berkeley Objects (ABO) \cite{collins2021abo}, a large-scale dataset containing nearly \textbf{8,000} real household objects from \textbf{98} categories with their corresponding nature language descriptions. We use ABO dataset because it contains the categories of other small datasets, such as ShapeNet\cite{chen2018text2shape}. Benefiting from their detailed texture and non-lambertian BRDFs, the 3D models in ABO can be photo-realistically rendered. To construct multi-view images dataset with their nature language descriptions, we use Blender \cite{blender2008} to render each 3D model into $256 \times 256$ RGB-alpha images from 36 cameras. Camera elevation is set as ${\rm{ - }}30^\circ$ and camera azimuth is sampled uniformly from the range $\left[ {{\rm{ - 180}}^\circ {\rm{, 180}}^\circ } \right]$. Totally, 286,308 multi-view images are rendered from 7,953 objects belong to 98 categories. 
We randomly split 80\%, 10\%, 10\% objects as our training, validation, and test set, respectively.

\textbf{Metrics.}
Following the settings of \cite{yu2021pixelnerf}, we evaluate the quality of our generated 3D objects by measuring the quality of novel view synthesis. Specifically, PSNR, SSIM\cite{wang2004image}, LPIPS \cite{zhang2018unreasonable} are adopted as our metrics. We also compute the CLIP score \cite{radford2021clip} between the rendered novel view images and the corresponding natural language description, which can measure the semantic consistency of the generated 3D objects with the description. The ResNet-50 CLIP model is adopted. Besides, we also adopt human evaluation for comparison. Two metrics are considered: \textbf{object fidelity} (including view fidelity and consistency among different views) and \textbf{caption similarity}. For each input caption, results from different methods are shown in random order and the workers are asked to order different results in terms of these two metrics. The average rank from different workers is used as the final score. 94 human evaluation results are collected. The higher PSNR, SSIM and CLIP score, the better; the lower LPIPS, object fidelity and caption similarity, the better.

\textbf{Experimental Setup.}
We implement our algorithm with Pytorch. The hyper-parameters of $\lambda_{pose}$, $\lambda_{txt}$, $\lambda_{prior}$, $\lambda_{img}$, $\lambda_{pixel}$, $\lambda_{caption}$ and $\lambda_{contrastive}$ are set to 0.1, 0.1, 0.1, 0.6, 1, 1 and 1 respectively.  
For our text-to-views generation module, we use AdamW optimizer to train 20 epochs.
For the views-to-3D generation module, we use Adam optimizer to train 100 epochs and randomly select $9$ views during each training step. More details are provided in the Appendix.

\begin{figure}[ht]
\centering
\includegraphics[width=0.9\linewidth]{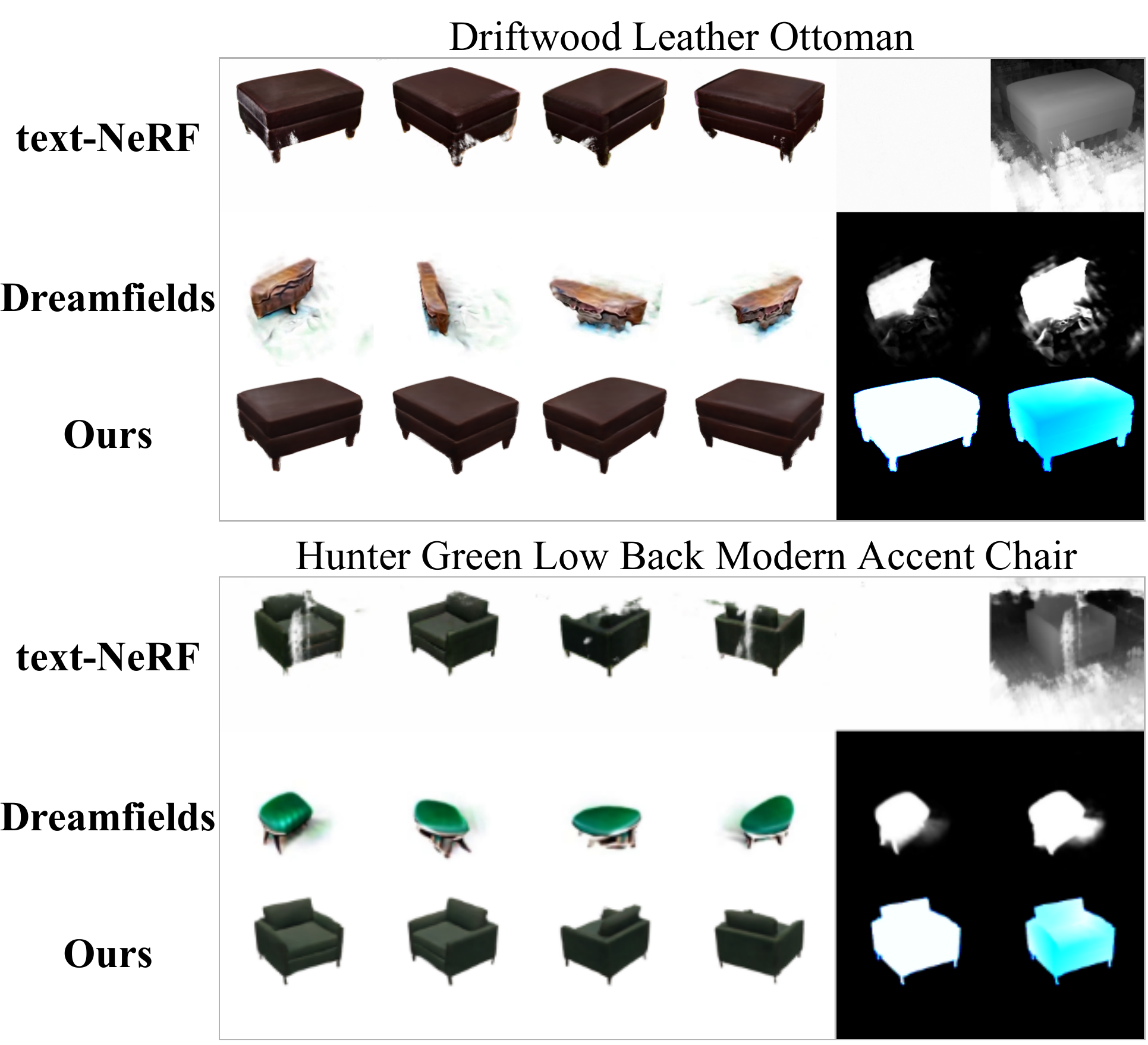}
\caption{Visual comparison against two baseline methods. For each sub figure, the textual title is the input caption, and the first 4 images are rendered novel views of the generated 3D object while the last two images are transmittance and depth from the first view respectively.}
\label{fig:compare_baseline}
\end{figure}

\subsection{Comparison Against Baselines}
As our 3D-TOGO model generates objects in the form of neural radiance fields, so we select two NeRF-based text-guided 3D object generation methods as our baseline: text-to-views generation module + NeRF \cite{mildenhall2020nerf} (called text-NeRF for convenient in the following) and Dreamfields \cite{jain2021dreamfields}. We use the code open-sourced by the authors.
As both text-NeRF and Dreamfields require training an individual network for each given natural language description, we randomly select $12$ text descriptions from our test set as the input captions. The selected text descriptions are included in the Appendix. As there is no ground truth for the generated views of Dreamfields, we do not use PSNR, SSIM and LPIPS in the comparison against Dreamfields.

\looseness=-1 Table \ref{table:compare_baseline} shows the quantitative comparison among different methods. Results of the first 6 descriptions and the average of 12 descriptions are shown while results of the rest 6 text descriptions are included in the Appendix. As we can see, for all cases, 3D-TOGO achieves the best results. Additionally, Figure \ref{fig:compare_baseline} shows the qualitative comparison among different methods. Results of the first 2 descriptions are shown while results of the rest 10 descriptions are included in the Appendix. As we can see, text-NeRF generates broken objects, as there are still some small inconsistent contents among the generated images.
Figure \ref{fig:compare_baseline} shows that Dreamfields cannot generate reasonable results. This may be because Dreamfields cannot generalize to household objects or it requires attentive hyperparameter-tuning. Besides, both text-NeRF and Dreamfields require time-consuming per-case optimization, while 3D-TOGO can be used across objects.

\subsection{Text-Guided 3D Object Generation
}

\begin{figure}[!t]
\centering
\includegraphics[width=0.9\linewidth]{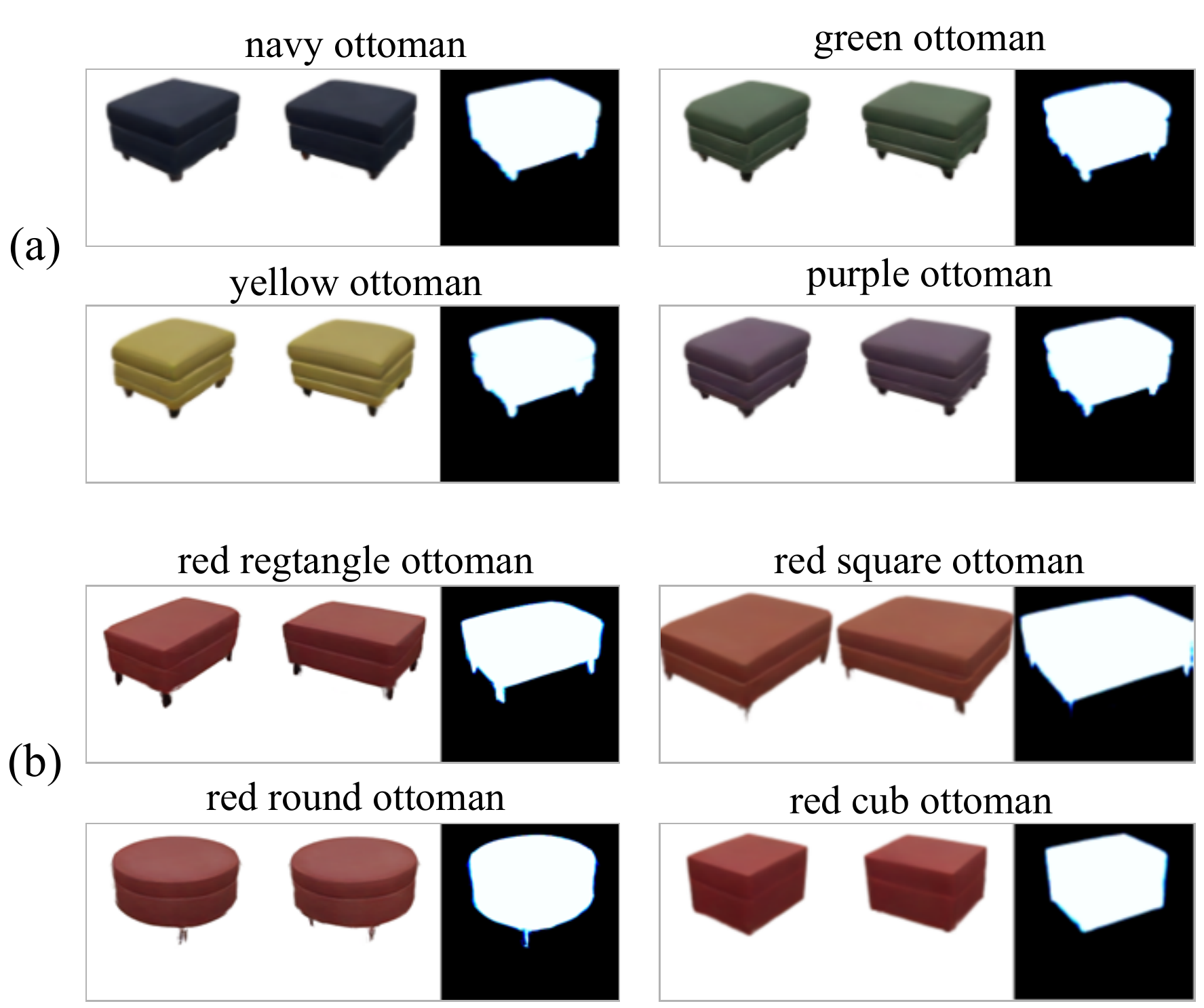}
\caption{3D object generation results with controlled (a) color and (b) shape. For each case, we show the input caption, 2 rendered novel views of the generated 3D object and the transmittance from the first view.
}
\label{fig:control_color_shape}
\vspace{-0.05in}
\end{figure}

Figure \ref{fig:main_results} shows the cross-category text-guided 3D generation results of our proposed 3D-TOGO model. Our model generates high-fidelity 3D objects matching the input caption across different object categories. Besides, Figure \ref{fig:control_color_shape} shows that our method can control the color and shape of the generated 3D objects by the input caption flexibly. More results are included in the Appendix.

\begin{table}
\footnotesize
\centering
\caption{\looseness=-1 Ablation study of our text-to-views generation module. `$prior$', `contrastive', `caption' and `L1' indicates $prior$-guidance, view contrastive learning, caption-guidance and  pixel-level L1 loss respectively.}
\label{tab:ablation-2d}
\begin{tabular}{llll|c}
\toprule [1pt]
$prior$ & contrastive & caption      & L1        & consistency-error $\downarrow$ \\ \midrule
      &             &           &           & 9.47              \\
\checkmark     &             &  &  & 8.88              \\
     &  \checkmark          & & & 9.17            \\
     &             & \checkmark &  & 9.23             \\
     &             & & \checkmark & 9.35             \\
\checkmark     & \checkmark           &           &           & 8.61              \\
\checkmark     &             & \checkmark         & \textbf{} & 8.81              \\
\checkmark     &             & \checkmark         & \checkmark         & 8.74              \\
\checkmark     & \checkmark           & \checkmark         & \checkmark         & \textbf{8.56}     \\ \hline\hline
\multicolumn{4}{c}{GT}                      & 7.55     \\
\bottomrule [1pt]
\end{tabular}
\vspace{-0.15in}
\end{table}

\subsection{Ablation Study}

In this section, we first study the effectiveness of different objectives on the quality of generated views from the text-to-views generation module and 3D objects from the views-to-3D generation module respectively. For quantitative comparison of the quality of generated views, we adopt metrics including FID \cite{heusel2017FID}, KID \cite{kid_score}, CLIP score and consistency error. 
Consistency error measures the average L2 error between views of adjacent camera poses and reflects view consistency to some degree. The lower the consistency error, the better view consistency. Results of consistency error are shown in Table \ref{tab:ablation-2d} while results of FID, KID, and CLIP-score are included in the Appendix. 
Then we study the effect of the number of views $N$ used for the views-to-3D generation module.

\textbf{View Generation Quality.} Table \ref{tab:ablation-2d} shows the quantitative results of our text-to-views generation module. As we can see, $prior$-guidance improves the consistency-error from 9.47 to 8.88, and view contrastive learning further improves it to 8.61, indicating both of these two improvements contribute to improving view-consistency. Besides, our complete text-to-views generation module achieves the best consistency-error of 8.56. 

\textbf{3D Object Generation Quality.}
Table \ref{tab:ablation-2d-3d} shows the CLIP-scores of our views-to-3D generation module for different object categories. The detailed results for each category will be provided in the Appendix. As we can see, 3D object generation based on the results of our complete text-to-views generation module achieves the best CLIP-score in all shown categories and achieves the best average CLIP-score among all categories of the ABO test set.

\begin{table}[t]
\caption{Ablation study of our text-to-views generation module and the effectiveness on 3D object generation.}
\footnotesize
\centering
\label{tab:ablation-2d-3d}
\begin{tabular}{llll|c}
\toprule [1pt]
\multicolumn{1}{c}{$prior$} & \multicolumn{1}{c}{contrastive} & \multicolumn{1}{c}{caption} & \multicolumn{1}{c|}{L1} & CLIP-score $\uparrow$ \\ \midrule
                          &                                 &                          &                         & 19.91                       \\
\checkmark                         &                                 &                          &                         &   20.50                       \\
& \checkmark & & & 19.97 \\    
& & \checkmark & & 20.14 \\
& & & \checkmark & 19.99 \\
\checkmark                         & \checkmark                               &                          &                            & 20.54                       \\
\checkmark                         &                                 & \checkmark                        &                           & 20.90                       \\
\checkmark                         &                                 & \checkmark                        & \checkmark                       & 20.93                       \\
\checkmark                         & \checkmark                               & \checkmark                        & \checkmark           &    \textbf{21.01}            \\ 
\bottomrule [1pt]
\end{tabular}
\end{table}

\begin{table}
\footnotesize
\centering
\caption{Ablation study results on the ABO test set regarding the number of views $N$.}
\label{tab:ablation-3d-num-view}
\resizebox{.43\textwidth}{!}{%
\begin{tabular}{c|cccc}
\toprule [1pt]
\# & PSNR $\uparrow$                & SSIM $\uparrow$                & LPIPS  $\downarrow$              & CLIP-score $\uparrow$          \\ \midrule
1                      & 9.69                 & 0.512               & 0.540               & 12.99                \\
3                      & 18.57                & 0.800               & 0.240               & 16.36                \\
6                      & 21.00                & 0.868               & 0.137               & 20.52                \\
9                      & 21.18                & 0.873               & \textbf{0.133}               & \textbf{20.71}                \\
18                     & \textbf{21.30} &  \textbf{0.877} & \textbf{0.133} & 20.54  \\
\bottomrule [1pt]
\end{tabular}
}
\end{table}

\textbf{Number of views $N$.}
Table \ref{tab:ablation-3d-num-view} shows the quantitative results of different number of views $N$ used for views-to-3D generation module. We evaluate the quality of generated 3D objects by measuring the quality of novel view synthesis. As we can see, more views yield better results in general. 9 views yield similar results to 18 views, so we set $N=9$ in our aforementioned experiments.

\section{Conclusion}
\label{sec:conclusion}
\looseness=-1
In this paper, we propose the 3D-TOGO model for the first attempt to achieve the generic text-guided 3D object generation. Our 3D-TOGO integrates a view-consistent text-to-views generation module for generating views of the target 3D object given an input caption; and a generic cross-scene neural rendering module for 3D object generation. For the text-to-views generation module, we adopt fine-grained pixel-level supervision signals, $prior$-guidance, caption-guidance and view contrastive learning for achieving better view fidelity, view-consistency and caption similarity.
A pixelNeRF model is adopted for the generic implicit 3D neural representation synthesis module. 
Extensive experiments on the largest 3D object dataset ABO show that our proposed 3D-TOGO model can better generate high-quality 3D objects according to the input captions across 98 different object categories both quantitatively and qualitatively, compared against text-NeRF and Dreamfields \cite{jain2021dreamfields}. Our 3D-TOGO model also allows for flexible control over categories, colors and shapes with the input caption. We describe the potential negative societal impacts and limitations of our work in the Appendix.


\bibliography{aaai23}

\begin{thebibliography}{55}
\providecommand{\natexlab}[1]{#1}

\bibitem[{Arjovsky, Chintala, and Bottou(2017)}]{arjovsky2017wasserstein}
Arjovsky, M.; Chintala, S.; and Bottou, L. 2017.
\newblock Wasserstein generative adversarial networks.
\newblock In \emph{International conference on machine learning}, 214--223.
  PMLR.

\bibitem[{Bengio, L{\'e}onard, and Courville(2013)}]{bengio2013estimating}
Bengio, Y.; L{\'e}onard, N.; and Courville, A. 2013.
\newblock Estimating or propagating gradients through stochastic neurons for
  conditional computation.
\newblock \emph{arXiv preprint arXiv:1308.3432}.

\bibitem[{Bi{\'n}kowski et~al.(2018)Bi{\'n}kowski, Sutherland, Arbel, and
  Gretton}]{kid_score}
Bi{\'n}kowski, M.; Sutherland, D.~J.; Arbel, M.; and Gretton, A. 2018.
\newblock Demystifying mmd gans.
\newblock \emph{arXiv preprint arXiv:1801.01401}.

\bibitem[{Canfes et~al.(2022)Canfes, Atasoy, Dirik, and
  Yanardag}]{canfes2022text}
Canfes, Z.; Atasoy, M.~F.; Dirik, A.; and Yanardag, P. 2022.
\newblock Text and Image Guided 3D Avatar Generation and Manipulation.
\newblock \emph{arXiv preprint arXiv:2202.06079}.

\bibitem[{Chen et~al.(2018)Chen, Choy, Savva, Chang, Funkhouser, and
  Savarese}]{chen2018text2shape}
Chen, K.; Choy, C.~B.; Savva, M.; Chang, A.~X.; Funkhouser, T.; and Savarese,
  S. 2018.
\newblock Text2shape: Generating shapes from natural language by learning joint
  embeddings.
\newblock In \emph{Asian conference on computer vision}, 100--116. Springer.

\bibitem[{Chen et~al.(2020)Chen, Li, Yu, El~Kholy, Ahmed, Gan, Cheng, and
  Liu}]{chen2020uniter}
Chen, Y.-C.; Li, L.; Yu, L.; El~Kholy, A.; Ahmed, F.; Gan, Z.; Cheng, Y.; and
  Liu, J. 2020.
\newblock Uniter: Universal image-text representation learning.
\newblock In \emph{European conference on computer vision}, 104--120. Springer.

\bibitem[{Collins et~al.(2022)Collins, Goel, Deng, Luthra, Xu, Gundogdu, Zhang,
  Vicente, Dideriksen, Arora et~al.}]{collins2021abo}
Collins, J.; Goel, S.; Deng, K.; Luthra, A.; Xu, L.; Gundogdu, E.; Zhang, X.;
  Vicente, T. F.~Y.; Dideriksen, T.; Arora, H.; et~al. 2022.
\newblock Abo: Dataset and benchmarks for real-world 3d object understanding.
\newblock In \emph{Proceedings of the IEEE/CVF Conference on Computer Vision
  and Pattern Recognition}, 21126--21136.

\bibitem[{Community(2018)}]{blender2008}
Community, B.~O. 2018.
\newblock Blender - a 3D modelling and rendering package.
\newblock \emph{Blender Foundation, Stichting Blender Foundation, Amsterdam}.

\bibitem[{Ding et~al.(2021)Ding, Yang, Hong, Zheng, Zhou, Yin, Lin, Zou, Shao,
  Yang et~al.}]{ding2021cogview}
Ding, M.; Yang, Z.; Hong, W.; Zheng, W.; Zhou, C.; Yin, D.; Lin, J.; Zou, X.;
  Shao, Z.; Yang, H.; et~al. 2021.
\newblock Cogview: Mastering text-to-image generation via transformers.
\newblock \emph{Advances in Neural Information Processing Systems}, 34:
  19822--19835.

\bibitem[{Ding et~al.(2022)Ding, Zheng, Hong, and Tang}]{ding2022cogview2}
Ding, M.; Zheng, W.; Hong, W.; and Tang, J. 2022.
\newblock CogView2: Faster and Better Text-to-Image Generation via Hierarchical
  Transformers.
\newblock \emph{arXiv preprint arXiv:2204.14217}.

\bibitem[{Dong et~al.(2017)Dong, Yu, Wu, and Guo}]{sisgan}
Dong, H.; Yu, S.; Wu, C.; and Guo, Y. 2017.
\newblock Semantic image synthesis via adversarial learning.
\newblock In \emph{Proceedings of the IEEE International Conference on Computer
  Vision}, 5706--5714.

\bibitem[{Esser et~al.(2021)Esser, Rombach, Blattmann, and
  Ommer}]{esser2021imagebart}
Esser, P.; Rombach, R.; Blattmann, A.; and Ommer, B. 2021.
\newblock Imagebart: Bidirectional context with multinomial diffusion for
  autoregressive image synthesis.
\newblock \emph{Advances in Neural Information Processing Systems}, 34:
  3518--3532.

\bibitem[{Esser, Rombach, and Ommer(2021)}]{vqgan}
Esser, P.; Rombach, R.; and Ommer, B. 2021.
\newblock Taming transformers for high-resolution image synthesis.
\newblock In \emph{Proceedings of the IEEE/CVF Conference on Computer Vision
  and Pattern Recognition}, 12873--12883.

\bibitem[{Goodfellow et~al.(2014)Goodfellow, Pouget-Abadie, Mirza, Xu,
  Warde-Farley, Ozair, Courville, and Bengio}]{goodfellow2014gan}
Goodfellow, I.; Pouget-Abadie, J.; Mirza, M.; Xu, B.; Warde-Farley, D.; Ozair,
  S.; Courville, A.; and Bengio, Y. 2014.
\newblock Generative adversarial nets.
\newblock \emph{Advances in neural information processing systems}, 27.

\bibitem[{Heusel et~al.(2017)Heusel, Ramsauer, Unterthiner, Nessler, and
  Hochreiter}]{heusel2017FID}
Heusel, M.; Ramsauer, H.; Unterthiner, T.; Nessler, B.; and Hochreiter, S.
  2017.
\newblock Gans trained by a two time-scale update rule converge to a local nash
  equilibrium.
\newblock \emph{Advances in neural information processing systems}, 30.

\bibitem[{Ho, Jain, and Abbeel(2020)}]{ho2020ddpm}
Ho, J.; Jain, A.; and Abbeel, P. 2020.
\newblock Denoising diffusion probabilistic models.
\newblock \emph{Advances in Neural Information Processing Systems}, 33:
  6840--6851.

\bibitem[{Hong et~al.(2022)Hong, Zhang, Pan, Cai, Yang, and
  Liu}]{hong2022avatarclip}
Hong, F.; Zhang, M.; Pan, L.; Cai, Z.; Yang, L.; and Liu, Z. 2022.
\newblock AvatarCLIP: Zero-Shot Text-Driven Generation and Animation of 3D
  Avatars.
\newblock \emph{ACM Transactions on Graphics (TOG)}, 41(4): 1--19.

\bibitem[{Hu et~al.(2021)Hu, Qi, Zhang, Pan, and Xu}]{hu2021text}
Hu, L.; Qi, J.; Zhang, B.; Pan, P.; and Xu, Y. 2021.
\newblock Text-driven 3D Avatar Animation with Emotional and Expressive
  Behaviors.
\newblock In \emph{Proceedings of the 29th ACM International Conference on
  Multimedia}, 2816--2818.

\bibitem[{Jain et~al.(2022)Jain, Mildenhall, Barron, Abbeel, and
  Poole}]{jain2021dreamfields}
Jain, A.; Mildenhall, B.; Barron, J.~T.; Abbeel, P.; and Poole, B. 2022.
\newblock Zero-shot text-guided object generation with dream fields.
\newblock In \emph{Proceedings of the IEEE/CVF Conference on Computer Vision
  and Pattern Recognition}, 867--876.

\bibitem[{Johnson, Alahi, and Fei-Fei(2016)}]{johnson2016perceptual}
Johnson, J.; Alahi, A.; and Fei-Fei, L. 2016.
\newblock Perceptual losses for real-time style transfer and super-resolution.
\newblock In \emph{European conference on computer vision}, 694--711. Springer.

\bibitem[{Khalid et~al.(2022{\natexlab{a}})Khalid, Xie, Belilovsky, and
  Popa}]{khalid2022text}
Khalid, N.; Xie, T.; Belilovsky, E.; and Popa, T. 2022{\natexlab{a}}.
\newblock Text to Mesh Without 3D Supervision Using Limit Subdivision.
\newblock \emph{arXiv preprint arXiv:2203.13333}.

\bibitem[{Khalid et~al.(2022{\natexlab{b}})Khalid, Xie, Belilovsky, and
  Tiberiu}]{khalid2022clip}
Khalid, N.~M.; Xie, T.; Belilovsky, E.; and Tiberiu, P. 2022{\natexlab{b}}.
\newblock Clip-mesh: Generating textured meshes from text using pretrained
  image-text models.
\newblock \emph{ACM Transactions on Graphics (TOG), Proc. SIGGRAPH Asia}.

\bibitem[{Lee et~al.(2022)Lee, Kim, Kim, Cho, and Han}]{lee2022rq-vae}
Lee, D.; Kim, C.; Kim, S.; Cho, M.; and Han, W.-S. 2022.
\newblock Autoregressive Image Generation using Residual Quantization.
\newblock In \emph{Proceedings of the IEEE/CVF Conference on Computer Vision
  and Pattern Recognition}, 11523--11532.

\bibitem[{Li et~al.(2019{\natexlab{a}})Li, Qi, Lukasiewicz, and
  Torr}]{li2019controllable}
Li, B.; Qi, X.; Lukasiewicz, T.; and Torr, P.~H. 2019{\natexlab{a}}.
\newblock Controllable text-to-image generation.
\newblock In \emph{Proceedings of the 33rd International Conference on Neural
  Information Processing Systems}, 2065--2075.

\bibitem[{Li et~al.(2019{\natexlab{b}})Li, Yatskar, Yin, Hsieh, and
  Chang}]{li2019visualbert}
Li, L.~H.; Yatskar, M.; Yin, D.; Hsieh, C.-J.; and Chang, K.-W.
  2019{\natexlab{b}}.
\newblock Visualbert: A simple and performant baseline for vision and language.
\newblock Preprint arXiv:1908.03557.

\bibitem[{Liu et~al.(2022)Liu, Wang, Qi, and
  Fu}]{liu2022implicit-text-3D-generation}
Liu, Z.; Wang, Y.; Qi, X.; and Fu, C.-W. 2022.
\newblock Towards Implicit Text-Guided 3D Shape Generation.
\newblock In \emph{Proceedings of the IEEE/CVF Conference on Computer Vision
  and Pattern Recognition}, 17896--17906.

\bibitem[{Michel et~al.(2022)Michel, Bar-On, Liu, Benaim, and
  Hanocka}]{text2mesh}
Michel, O.; Bar-On, R.; Liu, R.; Benaim, S.; and Hanocka, R. 2022.
\newblock Text2mesh: Text-driven neural stylization for meshes.
\newblock In \emph{Proceedings of the IEEE/CVF Conference on Computer Vision
  and Pattern Recognition}, 13492--13502.

\bibitem[{Mildenhall et~al.(2020)Mildenhall, Srinivasan, Tancik, Barron,
  Ramamoorthi, and Ng}]{mildenhall2020nerf}
Mildenhall, B.; Srinivasan, P.~P.; Tancik, M.; Barron, J.~T.; Ramamoorthi, R.;
  and Ng, R. 2020.
\newblock Nerf: Representing scenes as neural radiance fields for view
  synthesis.
\newblock In \emph{European conference on computer vision}, 405--421. Springer.

\bibitem[{Ni et~al.(2021)Ni, Huang, Su, Cui, Bharti, Wang, Zhang, and
  Duan}]{ni2021m3p}
Ni, M.; Huang, H.; Su, L.; Cui, E.; Bharti, T.; Wang, L.; Zhang, D.; and Duan,
  N. 2021.
\newblock M3p: Learning universal representations via multitask multilingual
  multimodal pre-training.
\newblock In \emph{Proceedings of the IEEE/CVF Conference on Computer Vision
  and Pattern Recognition}, 3977--3986.

\bibitem[{Nichol et~al.(2021)Nichol, Dhariwal, Ramesh, Shyam, Mishkin, McGrew,
  Sutskever, and Chen}]{nichol2021glide}
Nichol, A.; Dhariwal, P.; Ramesh, A.; Shyam, P.; Mishkin, P.; McGrew, B.;
  Sutskever, I.; and Chen, M. 2021.
\newblock Glide: Towards photorealistic image generation and editing with
  text-guided diffusion models.
\newblock \emph{arXiv preprint arXiv:2112.10741}.

\bibitem[{Poole et~al.(2022)Poole, Jain, Barron, and
  Mildenhall}]{poole2022dreamfusion}
Poole, B.; Jain, A.; Barron, J.~T.; and Mildenhall, B. 2022.
\newblock Dreamfusion: Text-to-3d using 2d diffusion.
\newblock \emph{arXiv preprint arXiv:2209.14988}.

\bibitem[{Radford et~al.(2021)Radford, Kim, Hallacy, Ramesh, Goh, Agarwal,
  Sastry, Askell, Mishkin, Clark et~al.}]{radford2021clip}
Radford, A.; Kim, J.~W.; Hallacy, C.; Ramesh, A.; Goh, G.; Agarwal, S.; Sastry,
  G.; Askell, A.; Mishkin, P.; Clark, J.; et~al. 2021.
\newblock Learning transferable visual models from natural language
  supervision.
\newblock In \emph{International Conference on Machine Learning}, 8748--8763.
  PMLR.

\bibitem[{Ramesh et~al.(2022)Ramesh, Dhariwal, Nichol, Chu, and
  Chen}]{ramesh2022dalle2}
Ramesh, A.; Dhariwal, P.; Nichol, A.; Chu, C.; and Chen, M. 2022.
\newblock Hierarchical text-conditional image generation with clip latents.
\newblock \emph{arXiv preprint arXiv:2204.06125}.

\bibitem[{Ramesh et~al.(2021)Ramesh, Pavlov, Goh, Gray, Voss, Radford, Chen,
  and Sutskever}]{dalle}
Ramesh, A.; Pavlov, M.; Goh, G.; Gray, S.; Voss, C.; Radford, A.; Chen, M.; and
  Sutskever, I. 2021.
\newblock Zero-shot text-to-image generation.
\newblock In \emph{International Conference on Machine Learning}, 8821--8831.
  PMLR.

\bibitem[{Reed et~al.(2016)Reed, Akata, Yan, Logeswaran, Schiele, and
  Lee}]{GAN-INT-CLS}
Reed, S.; Akata, Z.; Yan, X.; Logeswaran, L.; Schiele, B.; and Lee, H. 2016.
\newblock Generative adversarial text to image synthesis.
\newblock In \emph{International Conference on Machine Learning}, 1060--1069.
  PMLR.

\bibitem[{Reizenstein et~al.(2021)Reizenstein, Shapovalov, Henzler, Sbordone,
  Labatut, and Novotny}]{reizenstein2021common}
Reizenstein, J.; Shapovalov, R.; Henzler, P.; Sbordone, L.; Labatut, P.; and
  Novotny, D. 2021.
\newblock Common objects in 3d: Large-scale learning and evaluation of
  real-life 3d category reconstruction.
\newblock In \emph{Proceedings of the IEEE/CVF International Conference on
  Computer Vision}, 10901--10911.

\bibitem[{Sanghi et~al.(2022)Sanghi, Chu, Lambourne, Wang, Cheng, Fumero, and
  Malekshan}]{sanghi2021clip-forge}
Sanghi, A.; Chu, H.; Lambourne, J.~G.; Wang, Y.; Cheng, C.-Y.; Fumero, M.; and
  Malekshan, K.~R. 2022.
\newblock Clip-forge: Towards zero-shot text-to-shape generation.
\newblock In \emph{Proceedings of the IEEE/CVF Conference on Computer Vision
  and Pattern Recognition}, 18603--18613.

\bibitem[{Schwarz et~al.(2020)Schwarz, Liao, Niemeyer, and
  Geiger}]{schwarz2020graf}
Schwarz, K.; Liao, Y.; Niemeyer, M.; and Geiger, A. 2020.
\newblock Graf: Generative radiance fields for 3d-aware image synthesis.
\newblock \emph{Advances in Neural Information Processing Systems}, 33:
  20154--20166.

\bibitem[{Sennrich, Haddow, and Birch(2015)}]{sennrich2015bpe}
Sennrich, R.; Haddow, B.; and Birch, A. 2015.
\newblock Neural machine translation of rare words with subword units.
\newblock \emph{arXiv preprint arXiv:1508.07909}.

\bibitem[{Tan and Bansal(2019)}]{tan2019lxmert}
Tan, H.; and Bansal, M. 2019.
\newblock Lxmert: Learning cross-modality encoder representations from
  transformers.
\newblock \emph{arXiv preprint arXiv:1908.07490}.

\bibitem[{Tao et~al.(2020)Tao, Tang, Wu, Sebe, Jing, Wu, and
  Bao}]{tao2020dfgan}
Tao, M.; Tang, H.; Wu, S.; Sebe, N.; Jing, X.-Y.; Wu, F.; and Bao, B. 2020.
\newblock Df-gan: Deep fusion generative adversarial networks for text-to-image
  synthesis.
\newblock \emph{arXiv preprint arXiv:2008.05865}.

\bibitem[{Vaswani et~al.(2017)Vaswani, Shazeer, Parmar, Uszkoreit, Jones,
  Gomez, Kaiser, and Polosukhin}]{vaswani2017attention}
Vaswani, A.; Shazeer, N.; Parmar, N.; Uszkoreit, J.; Jones, L.; Gomez, A.~N.;
  Kaiser, {\L}.; and Polosukhin, I. 2017.
\newblock Attention is all you need.
\newblock In \emph{Advances in neural information processing systems},
  5998--6008.

\bibitem[{Wang et~al.(2022)Wang, Chai, He, Chen, and Liao}]{wang2021clip}
Wang, C.; Chai, M.; He, M.; Chen, D.; and Liao, J. 2022.
\newblock Clip-nerf: Text-and-image driven manipulation of neural radiance
  fields.
\newblock In \emph{Proceedings of the IEEE/CVF Conference on Computer Vision
  and Pattern Recognition}, 3835--3844.

\bibitem[{Wang et~al.(2004)Wang, Bovik, Sheikh, and Simoncelli}]{wang2004image}
Wang, Z.; Bovik, A.~C.; Sheikh, H.~R.; and Simoncelli, E.~P. 2004.
\newblock Image quality assessment: from error visibility to structural
  similarity.
\newblock \emph{IEEE transactions on image processing}, 13(4): 600--612.

\bibitem[{Wang et~al.(2021)Wang, Yu, Yu, Dai, Tsvetkov, and
  Cao}]{wang2021simvlm}
Wang, Z.; Yu, J.; Yu, A.~W.; Dai, Z.; Tsvetkov, Y.; and Cao, Y. 2021.
\newblock SimVLM: Simple Visual Language Model Pretraining with Weak
  Supervision.
\newblock Preprint arXiv:2108.10904.

\bibitem[{Xu et~al.(2018)Xu, Zhang, Huang, Zhang, Gan, Huang, and
  He}]{xu2018attngan}
Xu, T.; Zhang, P.; Huang, Q.; Zhang, H.; Gan, Z.; Huang, X.; and He, X. 2018.
\newblock Attngan: Fine-grained text to image generation with attentional
  generative adversarial networks.
\newblock In \emph{Proceedings of the IEEE conference on computer vision and
  pattern recognition}, 1316--1324.

\bibitem[{Ye et~al.(2021)Ye, Yang, Takac, Sunderraman, and Ji}]{ye2021xmcgan}
Ye, H.; Yang, X.; Takac, M.; Sunderraman, R.; and Ji, S. 2021.
\newblock Improving text-to-image synthesis using contrastive learning.
\newblock \emph{arXiv preprint arXiv:2107.02423}.

\bibitem[{Yu et~al.(2021)Yu, Ye, Tancik, and Kanazawa}]{yu2021pixelnerf}
Yu, A.; Ye, V.; Tancik, M.; and Kanazawa, A. 2021.
\newblock pixelnerf: Neural radiance fields from one or few images.
\newblock In \emph{Proceedings of the IEEE/CVF Conference on Computer Vision
  and Pattern Recognition}, 4578--4587.

\bibitem[{Zhang et~al.(2017)Zhang, Xu, Li, Zhang, Wang, Huang, and
  Metaxas}]{zhang2017stackgan}
Zhang, H.; Xu, T.; Li, H.; Zhang, S.; Wang, X.; Huang, X.; and Metaxas, D.~N.
  2017.
\newblock Stackgan: Text to photo-realistic image synthesis with stacked
  generative adversarial networks.
\newblock In \emph{Proceedings of the IEEE international conference on computer
  vision}, 5907--5915.

\bibitem[{Zhang et~al.(2018{\natexlab{a}})Zhang, Xu, Li, Zhang, Wang, Huang,
  and Metaxas}]{zhang2018stackgan++}
Zhang, H.; Xu, T.; Li, H.; Zhang, S.; Wang, X.; Huang, X.; and Metaxas, D.~N.
  2018{\natexlab{a}}.
\newblock Stackgan++: Realistic image synthesis with stacked generative
  adversarial networks.
\newblock \emph{IEEE transactions on pattern analysis and machine
  intelligence}, 41(8): 1947--1962.

\bibitem[{Zhang et~al.(2018{\natexlab{b}})Zhang, Isola, Efros, Shechtman, and
  Wang}]{zhang2018unreasonable}
Zhang, R.; Isola, P.; Efros, A.~A.; Shechtman, E.; and Wang, O.
  2018{\natexlab{b}}.
\newblock The unreasonable effectiveness of deep features as a perceptual
  metric.
\newblock In \emph{Proceedings of the IEEE conference on computer vision and
  pattern recognition}, 586--595.

\bibitem[{Zhang et~al.(2021)Zhang, Ma, Zhou, Men, Li, Ding, Tang, Zhou, and
  Yang}]{zhang2021m6-ufc}
Zhang, Z.; Ma, J.; Zhou, C.; Men, R.; Li, Z.; Ding, M.; Tang, J.; Zhou, J.; and
  Yang, H. 2021.
\newblock M6-UFC: Unifying Multi-Modal Controls for Conditional Image
  Synthesis.
\newblock \emph{arXiv preprint arXiv:2105.14211}.

\bibitem[{Zhou, Du, and Wu(2021)}]{zhou20213d}
Zhou, L.; Du, Y.; and Wu, J. 2021.
\newblock 3d shape generation and completion through point-voxel diffusion.
\newblock In \emph{Proceedings of the IEEE/CVF International Conference on
  Computer Vision}, 5826--5835.

\bibitem[{Zhou et~al.(2021)Zhou, Zhou, Wang, Cheng, Li, Yu, and
  Liu}]{zhou2021uc2}
Zhou, M.; Zhou, L.; Wang, S.; Cheng, Y.; Li, L.; Yu, Z.; and Liu, J. 2021.
\newblock UC2: Universal Cross-lingual Cross-modal Vision-and-Language
  Pre-training.
\newblock In \emph{Proceedings of the IEEE/CVF Conference on Computer Vision
  and Pattern Recognition}, 4155--4165.

\bibitem[{Zhuge et~al.(2021)Zhuge, Gao, Fan, Jin, Chen, Zhou, Qiu, and
  Shao}]{zhuge2021kaleido}
Zhuge, M.; Gao, D.; Fan, D.-P.; Jin, L.; Chen, B.; Zhou, H.; Qiu, M.; and Shao,
  L. 2021.
\newblock Kaleido-BERT: Vision-Language Pre-training on Fashion Domain.
\newblock In \emph{Proceedings of the IEEE/CVF Conference on Computer Vision
  and Pattern Recognition}, 12647--12657.

\end{thebibliography}

\clearpage


\begin{appendices}

\section{Experimental Settings}
\subsection{Optimization Details}

We use the ABO \cite{collins2021abo} dataset for our experiments. We select items with English textual descriptions and 3D models. The original annotated descriptions do not look like natural language, so we use hand-crafted rules (including removing brand, removing number indicating object size, and moving words of color to the beginning of the sentence) to parse and transform them as close as possible to natural language.

\begin{figure}[H]
    \centering
    \begin{subfigure}[]{0.37\textwidth}
    \includegraphics[width=\textwidth]{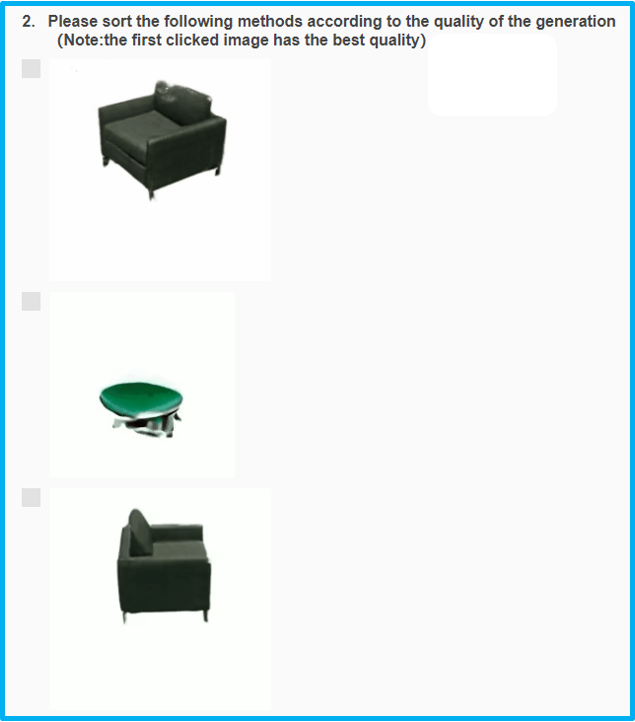}
    \caption{Screenshot for ordering according to object fidelity.}
    \end{subfigure}
    \begin{subfigure}[]{0.37\textwidth}
    \includegraphics[width=\textwidth]{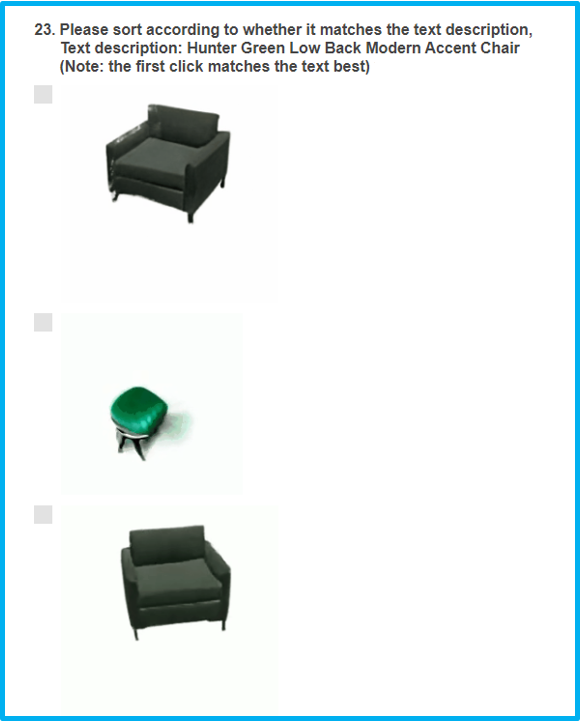}
    \caption{Screenshot for ordering according to caption similarity.}
    \end{subfigure}
    \caption{Screenshots of our volunteer questionnaire survey.}
\label{app-fig:human_study}
\end{figure}

We implement our algorithm with Pytorch. All experiments are conducted on servers with 8 Nvidia V100 GPU (32GB) cards and Intel Xeon Platinum 8168 CPU (2.70GHz). Hyperparameters are searched on the validation set.

For our text-to-views generation module, we use the released VQGAN \cite{vqgan} model as the image tokenizer. The size of the codebook is 1024. The input image size is $256 \times 256$ and the image tokenizer transforms the input image as $16 \times 16{\rm{ = }}256$ tokens. The vocabulary size of the BPE tokenizer is 49,408. The maximum length of the caption token sequence is 128. 36 camera poses across different objects are selected. We use a 24-layer Transformer with 8 attention heads. The dimension of each attention head is 64. We use the AdamW optimizer with $\beta_1=0.9$, $\beta_2=0.96$ to train 20 epochs with batch size 768. The initial learning rate is set as 1e-3 and decayed with a cosine annealing scheduler. The whole optimization process takes about 12 hours with 32 Nvidia V100 GPUs (32 GB).

\begin{table*}[htb]
\footnotesize
\centering
\caption{Text descriptions used for comparison against baselines}
\begin{tabular}{c|c}
\toprule[1pt]
TextId & Text descriptions  \\
\midrule
text1  & Driftwood Leather Ottoman                               \\
text2  & Hunter Green Low Back Modern Accent Chair                      \\
text3  & Cream Herringbone Queen Headboard             \\
text4  & Tan Tufted Sofa                                            \\
text5  & Light Grey with White Whipstitch Edge Decorative Throw Pillow \\
text6  & Blue Geometric Area Rug                                          \\
text7  & Passion Pink Old World Vintage Persian Area Rug                         \\
text8  & Charcoal Leather Loveseat                         \\
text9 & Linen Fabric Slipcover Sofa Couch                       \\
text10 & Pewter Marin Studded Sofa                                       \\
text11 & Ivory, Grey French Laundry Stripe Decorative Throw Pillow       \\
text12 & Platinum Black Mid-Century Tufted Customizable Daybed Sofa \\       
\bottomrule [1pt]
\end{tabular}
\label{table:compare_text_details}
\end{table*}

\begin{table*}[]
\footnotesize
\centering
\caption{
Quantitative comparison against text-NeRF (short for text-to-views generation + NeRF\cite{mildenhall2020nerf})   and Dreamfields \cite{jain2021dreamfields}.}
\begin{tabular}{c|c|ccccccc}
\toprule[1pt]
Metric                              & Method      & text7                             & text8                             & text9                             & text10                             & text11                             & text12                             & 12 Texts Avg.  \\ \midrule
\multirow{2}{*}{PSNR $\uparrow$}               & text-NeRF   & 16.33                             & 19.56                             & 20.09                              & 19.17                             & 13.28                             & 0.58                              & 14.04          \\
                                    & Ours        & \textbf{28.74}                    & \textbf{23.81}                    & \textbf{26.15}                    & \textbf{20.71}                    & \textbf{26.63}                    & \textbf{29.45}                    & \textbf{24.98} \\ \midrule
\multirow{2}{*}{SSIM $\uparrow$}               & text-NeRF   & 0.843                             & 0.866                             & 0.887                             & 0.839                             & 0.702                             & 0.001                             & 0.636          \\
                                    & Ours        & \textbf{0.873}                    & \textbf{0.897}                    & \textbf{0.928}                    & \textbf{0.865}                    & \textbf{0.889}                    & \textbf{0.952}                    & \textbf{0.900} \\ \midrule
\multirow{2}{*}{LPIPS $\downarrow$}              & text-NeRF   & 0.138                             & 0.110                             & 0.109                             & 0.137                             & 0.379                             & 0.482                             & 0.239          \\
                                    & Ours        & \textbf{0.099}                    & \textbf{0.015}                    & \textbf{0.084}                    & \textbf{0.122}                    & \textbf{0.092}                    & \textbf{0.062}                    & \textbf{0.092} \\ \midrule
\multirow{3}{*}{CLIP- Score $\uparrow$}        & Dreamfields & 18.63                             & 18.16                             & 16.39                             & 19.44                             & 20.11                             & 16.56                             & 18.40          \\
                                    & text-NeRF   & 18.05                             & \textbf{24.70}                             & 13.97                             & 22.22                             & 13.03                             & 11.61                             & 18.38          \\
                                    & Ours        & \textbf{24.63}                    & 24.12                    & \textbf{15.71}                    & \textbf{22.47}                    & \textbf{23.00}                    & \textbf{21.42}                    & \textbf{22.84} \\ \midrule
\multirow{3}{*}{Object Fidelity $\downarrow$}    & Dreamfields & 2.97 &   2.95     &    3.00     & 2.83     &  2.38       &    1.99      & 2.63           \\
                                    & text-NeRF   &     1.94     &  \textbf{1.52}       &    1.97       &    2.08   &    2.56      &     2.94      & 2.27           \\
                                    & Ours        & \textbf{1.09} & 1.54 & \textbf{1.03} & \textbf{1.08} & \textbf{1.05} & \textbf{1.07} & \textbf{1.11}  \\ \midrule
\multirow{3}{*}{Caption Similarity $\downarrow$} & Dreamfields &   2.96     &  2.91   & 2.92       &     2.87      &    2.47      &    2.02      & 2.64           \\
                                    & text-NeRF   &   1.91       &  \textbf{1.51}       &  2.04        &    2.07       &    2.49      &     2.94     & 2.25           \\
                                    & Ours        & \textbf{1.13} & 1.57 & \textbf{1.04} & \textbf{1.05} & \textbf{1.03} & \textbf{1.04} & \textbf{1.11} 
                                    \\
\bottomrule [1pt]
\end{tabular}
\label{app-table:compare_baseline}
\end{table*}

For our views-to-3D generation module, the input image size is $256 \times 256$. We use the Adam optimizer to train 100 epochs with the learning rate 1e-4 and randomly select 9 out of 36 views for training at each training step. During the training stage, We use a batch size of 8 objects and 128 rays per object. To avoid sampling invalid rays and improve optimization efficiency, we sample rays in the bounding box surrounding the object for the first 20 epochs. The bounding box is removed from 20 to 100 epochs to avoid background artifacts. Our views-to-3D generation module takes almost 1 day to train on 8 Nvidia V100 GPUs (32GB).


\begin{table*}[htbp]
\footnotesize
\centering
\caption{Ablation study of our 2D view-consistent text-to-image generation module. `$prior$' indicates $prior$-guidance, `contrastive' indicated view contrastive learning, `caption' indicates caption-guidance and `L1' indicates pixel-level L1 loss.}
\label{app-tab:ablation-2d}
\begin{tabular}{llll|cccc}
\toprule [1pt]
$prior$ & contrastive & caption & L1 & FID $\downarrow$   & KID (* 1e-3) $\downarrow$ & CLIP-score $\uparrow$ & consistency-error $\downarrow$ \\ \midrule
      &             &      &    & 15.72 & 3.76       & 20.58      & 9.47              \\
\checkmark     &             &      &    & 15.44 & 3.29       & 20.62      & 8.88              \\
     & \checkmark            &      &    & 15.13 & 2.95       & 20.67      & 9.17              \\
     &             &  \checkmark    &    & 14.43 & 2.80       & 21.23      & 9.23              \\
     &             &      & \checkmark   & 15.52 & 3.63       & 20.59      & 9.35              \\
\checkmark     & \checkmark           &      &    & 15.43 & 2.99       & 20.63      & 8.61              \\
\checkmark     &             & \checkmark    &    & 14.99 & 2.93       & \textbf{21.20}      & 8.81              \\
\checkmark     &             & \checkmark    & \checkmark  & \textbf{14.70} & 2.77       & \textbf{21.20}      & 8.74              \\
\checkmark     & \checkmark           & \checkmark    & \checkmark  & \textbf{14.70} & \textbf{2.64}       & \textbf{21.20}      & \textbf{8.56}              \\ \midrule
\multicolumn{4}{c|}{GT}         & 0     & 0          & 22.88      & 7.55              \\ 
\bottomrule [1pt]
\end{tabular}
\end{table*}

\begin{table*}[htbp]
\setlength\tabcolsep{3pt}
\vspace{-1mm}
\caption{Ablation study of our text-to-views generation module and the effectiveness on 3D object generation measured by CLIP-score. `$prior$', `contrastive', `caption' and `L1' indicates $prior$-guidance, view contrastive learning, caption-guidance and  pixel-level L1 loss respectively.}
\footnotesize
\centering
\label{tab:app-ablation-2d-3d}
\begin{tabular}{llll|ccccc}
\toprule [1pt]
\multicolumn{1}{c}{$prior$} & \multicolumn{1}{c}{contrastive} & \multicolumn{1}{c}{caption} & \multicolumn{1}{c|}{L1} & Chair          & Sofa           & Lamp           & Planter        & Pillow        \\ \midrule
                          &                                 &                          &                         & 20.44          & 20.25          & 18.32          & 23.57          & 21.21                            \\
\checkmark                         &                                 &                          &                         & 20.94          & 21.03          & 19.06          & 23.87          & 21.48                           \\
& \checkmark & & & 20.38 & 19.91 & 18.73 & 23.65& 20.90 \\    
& & \checkmark & & 20.78 & 20.36 & 19.12 & 23.96 & 21.43  \\
& & & \checkmark & 20.51 & 20.31 & 18.70 & 23.50 & 21.50 \\
\checkmark                         & \checkmark                               &                          &                         & 21.09          & 20.94          & 19.71          & 23.76          & 21.65                      \\
\checkmark                         &                                 & \checkmark                        &                         & 21.60          & 21.29          & 19.86          & 24.02          & 22.13                       \\
\checkmark                         &                                 & \checkmark                        & \checkmark                       & 21.57          & 21.46          & 19.20          & 23.91          & 21.90                           \\
\checkmark                         & \checkmark                               & \checkmark                        & \checkmark                       & \textbf{21.75} & \textbf{21.51} & \textbf{20.38} & \textbf{24.14} & \textbf{21.97}              \\ \midrule
\multicolumn{1}{c}{$prior$} & \multicolumn{1}{c}{contrastive} & \multicolumn{1}{c}{caption} & \multicolumn{1}{c|}{L1} & Tools          & Bed           & Dehumidifier           &    Light Fixture   &  Exercise Mat      \\ \midrule
                          &                                 &                          &                         & 15.12          & 19.60          & 17.62          &   15.79  & 21.77 \\
\checkmark                         &                                 &                          &                         & 15.34          &  19.94   & 17.33          &   16.00    & 22.69 \\
& \checkmark & & & 15.09 & 18.94 & 17.83 & 16.40& 22.40  \\    
& & \checkmark & & 15.45 & 19.05 & 18.23 & 16.85 & 23.08  \\
& & & \checkmark & 15.40 & 19.50 & 16.88 & 15.59 & 22.11  \\
\checkmark                         & \checkmark                               &                          &                         & 16.73          & 20.24            & 17.48          &  16.33       & 22.80  \\
\checkmark                         &                                 & \checkmark                        &                         & 15.19          & 20.48           & 18.86          &  16.82        & 23.51   \\
\checkmark                         &                                 & \checkmark                        & \checkmark                       & 15.43          & 20.21            & 18.85          &   15.61      & 23.32   \\
\checkmark                         & \checkmark                               & \checkmark                        & \checkmark                       & \textbf{19.28} & \textbf{20.17} & \textbf{19.77} & \textbf{17.29}  & \textbf{23.97} \\ \midrule
\multicolumn{1}{c}{$prior$} & \multicolumn{1}{c}{contrastive} & \multicolumn{1}{c}{caption} & \multicolumn{1}{c|}{L1} &  Home         & Instrument           &  Healthcare        &  Multiport Hub      & Recording Equipment \\ \midrule
                          &                                 &                          &                         & 17.10                    & 15.07                          & 21.96      & 16.88                             & 14.38                       \\
\checkmark                         &                                 &                          &                         &  17.55                    & 16.07                          & 22.60      & 17.19                             & 16.74                       \\
& \checkmark & & & 17.43 & 15.02 & 21.73 & 16.83& 16.62  \\    
& & \checkmark & & 17.14 & 13.78 & 23.54 & 17.00 & 16.90  \\
& & & \checkmark & 17.28 & 15.21 & 21.99 & 16.85 & 15.46  \\
\checkmark                         & \checkmark                               &                          &                         & 17.53                    & 16.35                          & 22.62      & 17.06                             & 16.34                       \\
\checkmark                         &                                 & \checkmark                        &                         &  17.58                    & 16.75                          & 23.17      & 16.94                             & 17.39                       \\
\checkmark                         &                                 & \checkmark                        & \checkmark                       &  17.48                    & 17.55                          & 23.90       & 17.23                             & 18.05                       \\
\checkmark                         & \checkmark                               & \checkmark                        & \checkmark                       & \textbf{18.10}           & \textbf{18.25}                 & \textbf{23.98}      & \textbf{17.39}                             & \textbf{18.67}              \\  \midrule
\multicolumn{1}{c}{$prior$} & \multicolumn{1}{c}{contrastive} & \multicolumn{1}{c}{caption} & \multicolumn{1}{c|}{L1} &  Ottoman         & Cabinet           &  Headboard        & Stool Seating      & \textbf{All Categiries Avg.} \\ \midrule
                          &                                 &                          &                         & 24.34          & 19.02          & 17.59  & 22.04 & 19.91                       \\
\checkmark                         &                                 &                          &                         &  25.35        & 20.05         &  17.62  & 22.39 & 20.50                       \\
& \checkmark & & & 24.28 & 19.23 & 17.43 & 21.96 & 19.79  \\    
& & \checkmark & & 24.58 & 19.26 & 17.99 & 22.39 & 20.14  \\
& & & \checkmark & 24.64 & 19.21 & 17.38 & 22.08 & 19.99  \\
\checkmark                         & \checkmark                               &                          &                         & 25.21           & 20.14          &   17.60  & 22.47   & 20.54                       \\
\checkmark                         &                                 & \checkmark                        &                         &  25.37         & 20.37         & 18.18   & 22.58 & 20.90                       \\
\checkmark                         &                                 & \checkmark                        & \checkmark                       &  25.48         & 20.24         & 18.17   & 22.59 & 20.93                       \\
\checkmark                         & \checkmark                               & \checkmark                        & \checkmark                       & \textbf{25.64}  & \textbf{20.61} & \textbf{18.26} & \textbf{22.86} & \textbf{21.01}              \\ 
\bottomrule [1pt]
\end{tabular}
\end{table*}

\subsection{Human Study}
Figure \ref{app-fig:human_study} shows the screenshots of our human study (volunteer questionnaire survey). The volunteers are asked to sort the results of different methods in terms of object fidelity (including view fidelity and consistency among different views) and caption similarity (semantic similarity between input caption and the generated 3D object). 94 effective questionnaires are collected eventually. The average rank of different volunteers is used as the final score.

\section{More Experimental Results}

\subsection{Comparison Against Baselines}

\begin{figure*}[htb]
\centering
\includegraphics[width=\linewidth]{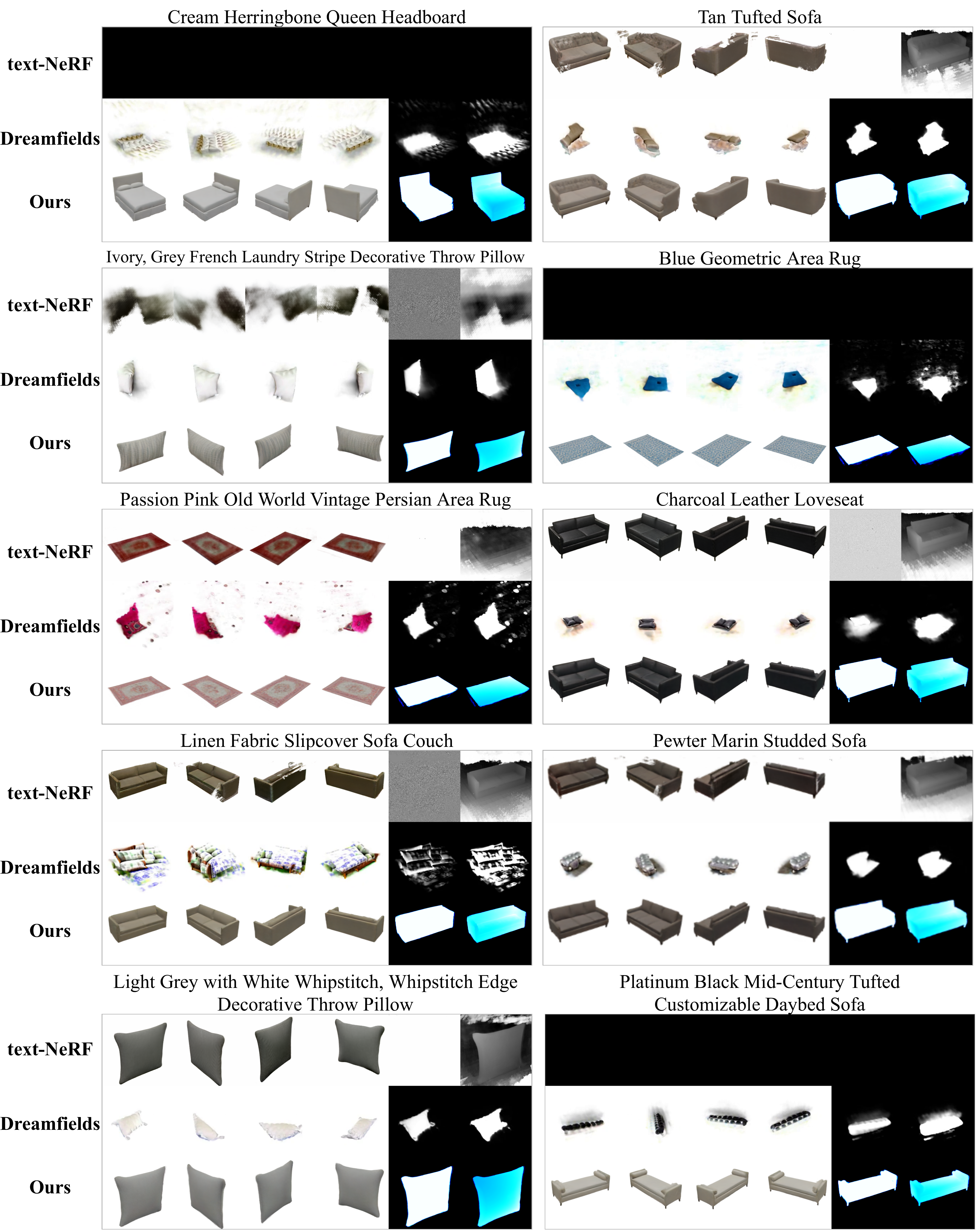}
\caption{Visual comparison against two baseline methods. For each subfigure, the textual title is the input caption, and the first 4 images are rendered novel views of the generated 3D object while the last two images are transmittance and depth from the first view respectively.}
\label{app-fig:compare_baseline}
\end{figure*}

We compare our 3D-TOGO model against two baselines: text-to-views generation module + NeRF \cite{mildenhall2020nerf} (called text-NeRF) and Dreamfields \cite{jain2021dreamfields}. As both text-NeRF and Dreamfields require training an individual network for each given natural language description, we randomly select $12$ text descriptions from our test set as the input captions.
For Dreamfields, we reduce the weight of transmittance loss, extend the duration of target transmittance annealing, and use the prompt engineering "a 3d render of furniture" for some input texts to get better results.
Table \ref{table:compare_text_details} lists the selected text descriptions.
In the main text, we show quantitative results of the first 6 texts, the results of the remaining 6 texts are shown in Table \ref{app-table:compare_baseline}. In the main text, we show qualitative results of the first 2 texts, the results of the rest 10 texts are shown in Figure \ref{app-fig:compare_baseline}.

\begin{figure*}[ht]
\centering
\includegraphics[width=\linewidth]{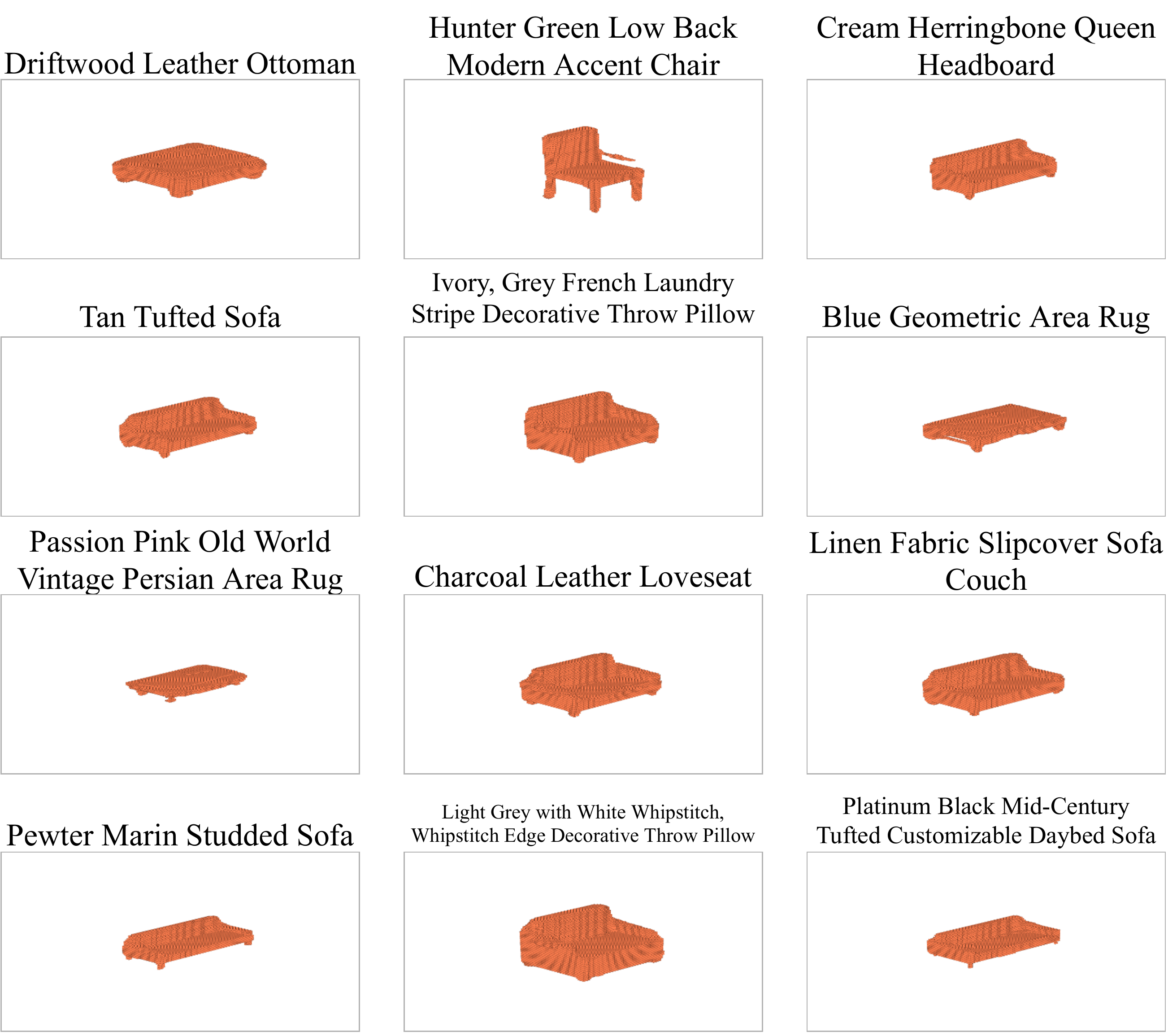}
\caption{Qualitative results of CLIP-forge\cite{sanghi2021clip-forge} on 12 texts in Table \ref{table:compare_text_details}.
}
\label{app-fig:clip_forge_12text}
\vspace{-0.15in}
\end{figure*}

We also compare our 3D-TOGO model against CLIP-Forge\cite{sanghi2021clip-forge}. As CLIP-Forge generates 3D objects without color and texture, we only conduct the qualitative comparison. Figure \ref{app-fig:clip_forge_12text} shows the qualitative results of CLIP-forge on the 12 text descriptions in Table \ref{table:compare_text_details}. As we can see, CLIP-forge can generate desired 3D shapes for ottoman, chair, loveseat, and sofa categories, while fails for pillow, headboard, and rug categories. Compared with CLIP-Forge, our 3D-TOGO can generate high-quality realistic 3D objects with color and rich textures on all these input captions.

\subsection{Text-Guided 3D Object Generation}
\begin{figure*}[!t]
\centering
\includegraphics[width=\linewidth]{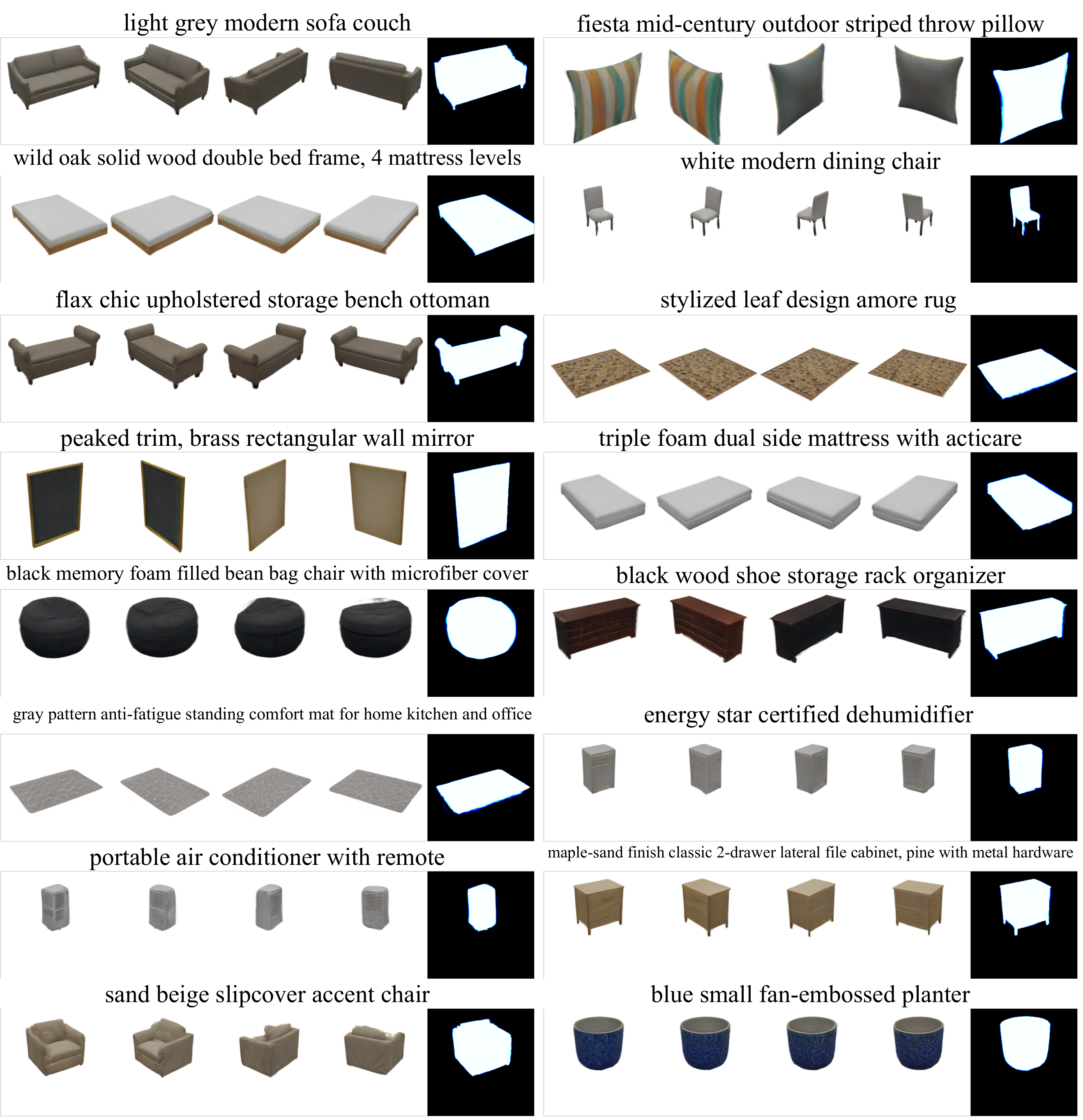}
\vspace{-0.1in}
\caption{Generation results of our proposed 3D-TOGO model. For each case, 
we show the input caption, 4 rendered novel views of the generated 3D object and the transmittance from the first view. 
}
\label{app-fig:main_result}
\vspace{-0.15in}
\end{figure*}

Figure \ref{app-fig:main_result} and Figure \ref{app-fig:main_result2} show more cross-category generation results of our 3D-TOGO model.
In the supplementary materials, we also provide explanatory supplementary videos of the generated text-matched 3D objects. These videos include renderings where the camera is
moved around the object.

\begin{figure*}[ht]
\centering
\includegraphics[width=\linewidth]{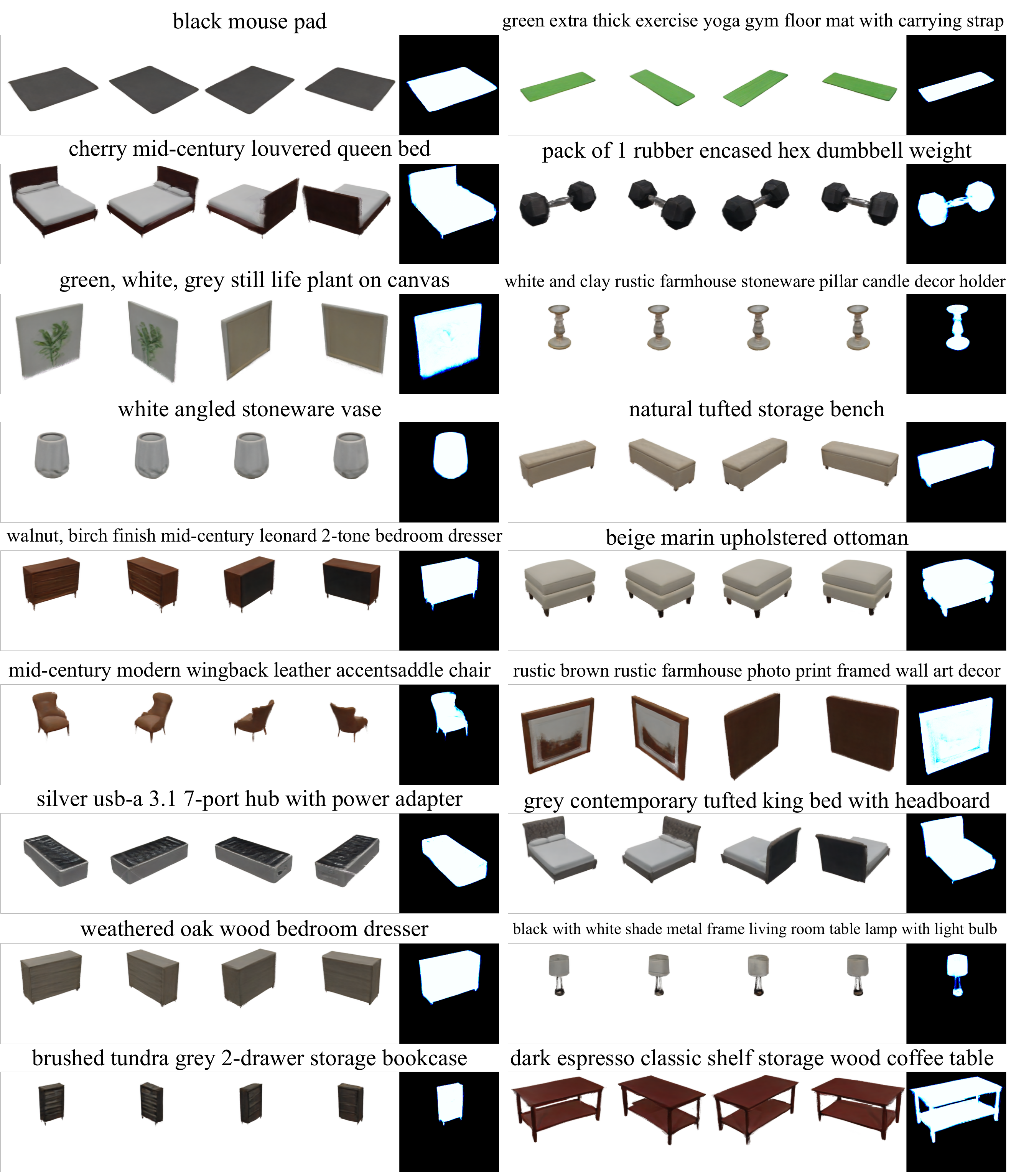}
\vspace{-0.1in}
\caption{Generation results of our proposed 3D-TOGO model. For each case, 
we show the input caption, 4 rendered novel views of the generated 3D object and the transmittance from the first view. 
}
\label{app-fig:main_result2}
\end{figure*}

\begin{figure*}[!t]
\centering
\includegraphics[width=1.\linewidth]{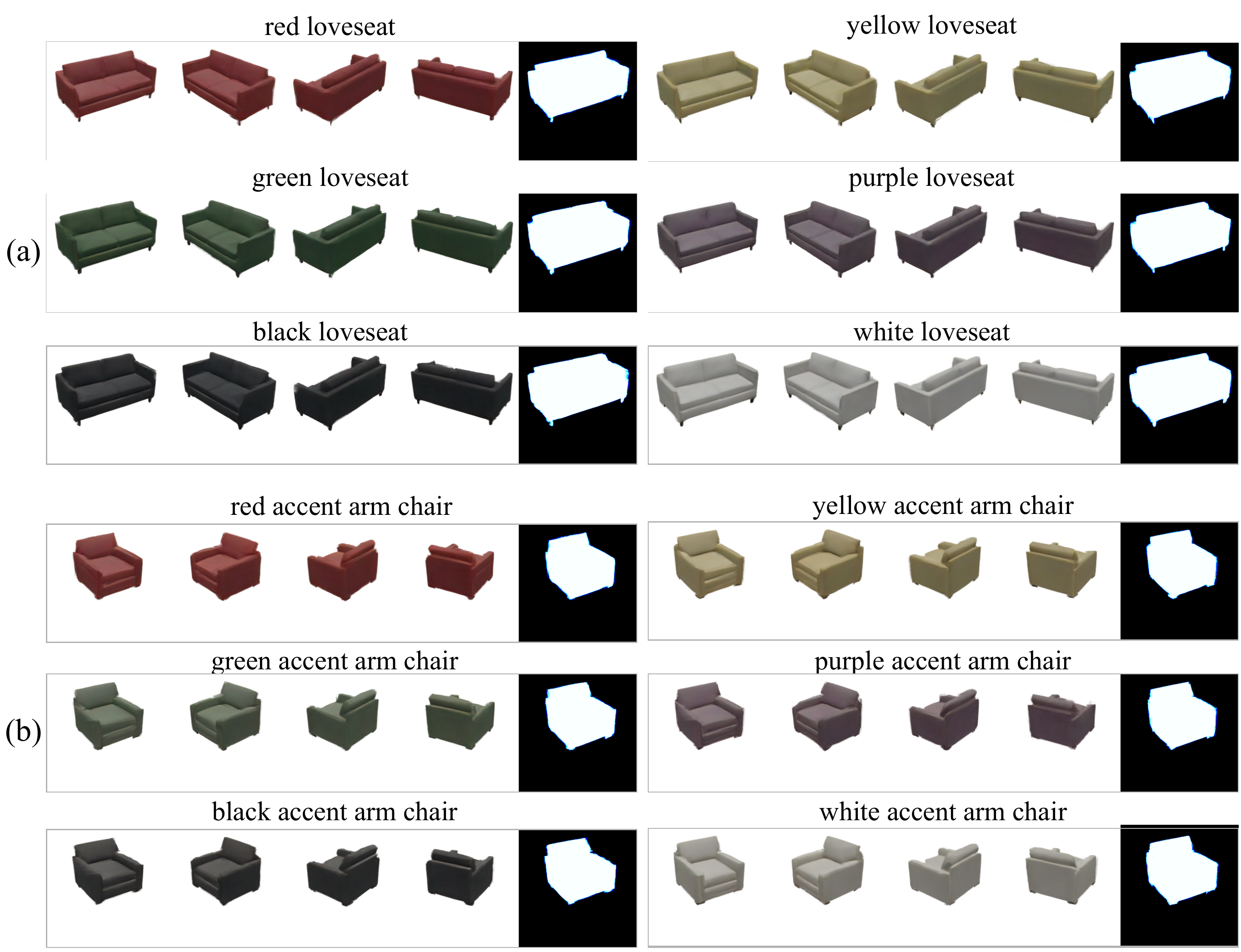}
\caption{3D object generation results with controlled color. For each case, we show the input caption, 4 rendered novel views of the generated 3D object and the transmittance from the first view.
}
\label{app-fig:control_color}
\end{figure*}

\begin{figure*}[!t]
\centering
\includegraphics[width=\linewidth]{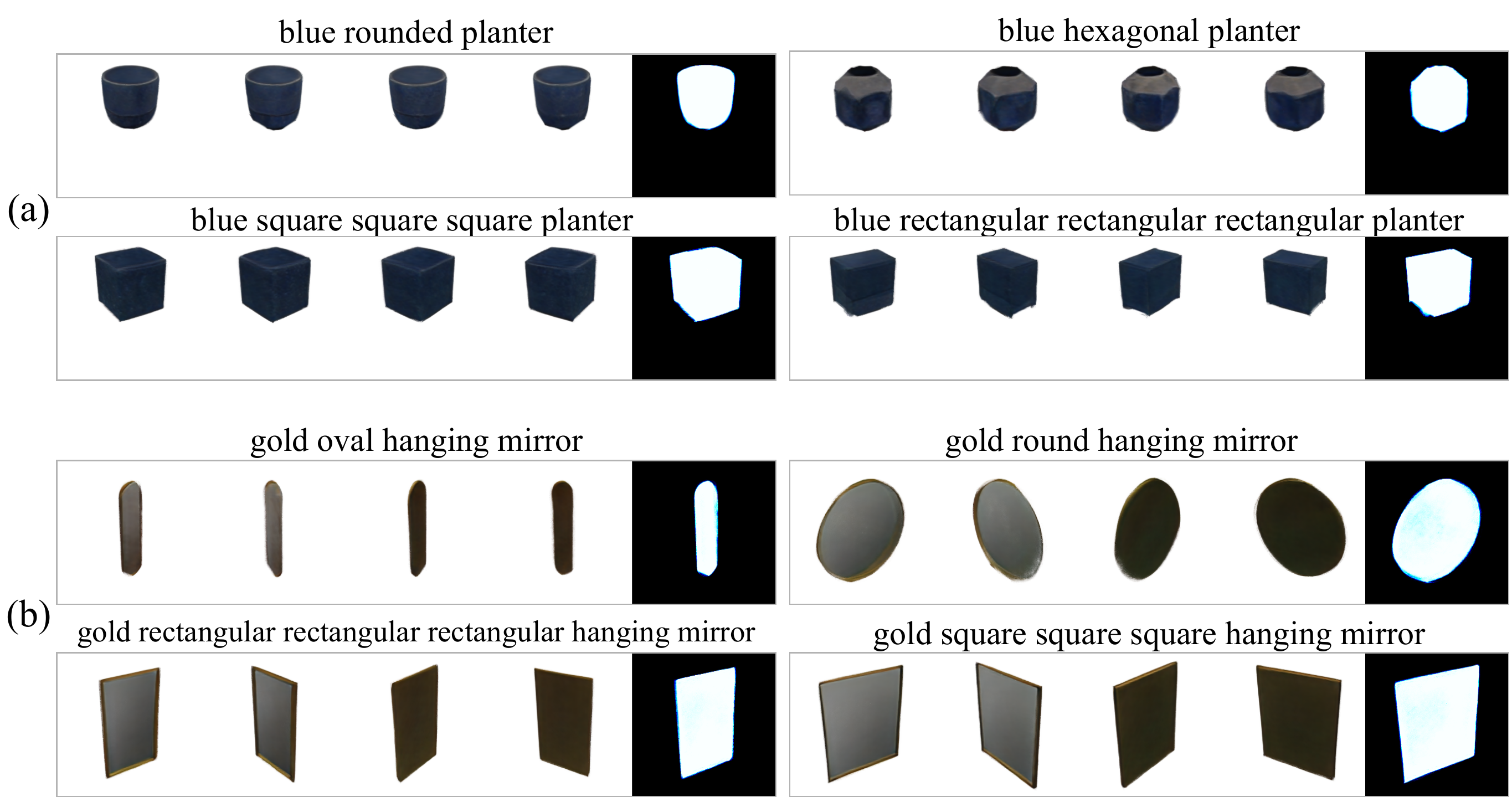}
\vspace{-0.2in}
\caption{3D object generation results with controlled shape. For each case, we show the input caption, 4 rendered novel views of the generated 3D object and the transmittance from the first view.
}
\label{app-fig:control_shape}
\end{figure*}

\begin{figure*}[ht]
\centering
\includegraphics[width=\linewidth]{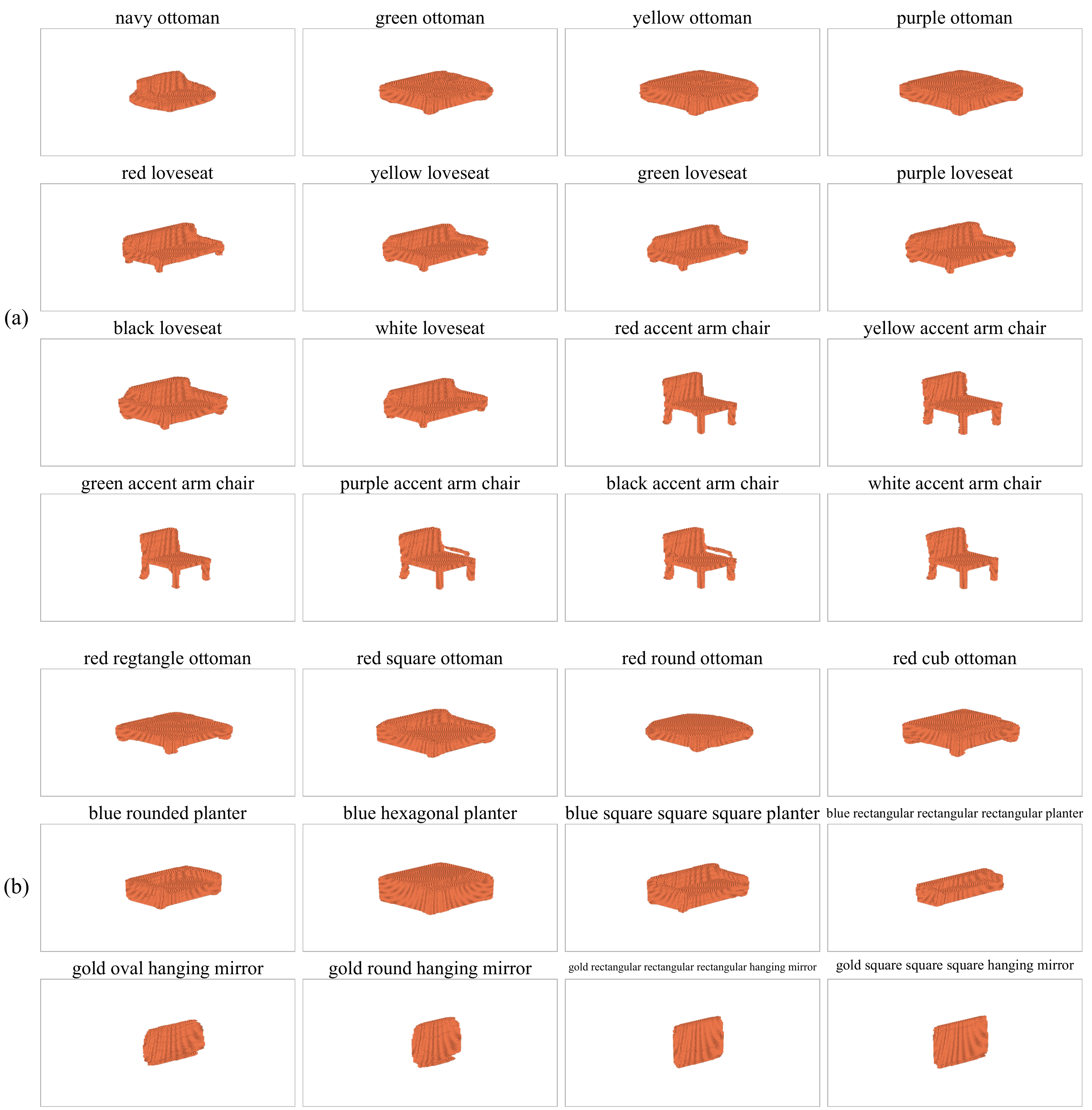}
\caption{Qualitative results of CLIP-forge\cite{sanghi2021clip-forge} for various color and shape.
}
\label{app-fig:clip_forge_colorshape}
\vspace{-0.15in}
\end{figure*}

To further demonstrate the superiority of our 3D-TOGO regarding the shape and color control, we conduct qualitative comparison experiments for 3D-TOGO and CLIP-Forge with changed color and shape. Figure \ref{app-fig:control_color} and Figure \ref{app-fig:control_shape} show more results of generating 3D objects with controlled color and shape respectively. Figure \ref{app-fig:clip_forge_colorshape} shows the qualitative results of CLIP-forge for various color and shape. As displayed in Figures \ref{app-fig:control_color}, \ref{app-fig:control_shape}, \ref{app-fig:clip_forge_colorshape},  CLIP-forge fails to control the color and shape of generated 3D objects, while our 3D-TOGO can generating realistic text-matched 3D objects.

\section{Ablation study}

As mentioned in the main text, in this section, we study the effectiveness of different objectives mentioned in the 3D-TOGO model on the quality of generated views from the text-to-views generation module and 3D objects from the views-to-3D generation module respectively. For quantitative comparison of the quality of generated views, we adopt metrics including FID \cite{heusel2017FID}, KID \cite{kid_score}, CLIP score and consistency error. 
FID and KID measure the distance between feature distributions of the real views and generated views. The lower FID and KID, the better view fidelity. CLIP score measures the semantic similarity between the input caption and the generated views and is the higher the better. 
Consistency error measures the average L2 error between views of adjacent camera poses and reflects view consistency to some degree. The lower the consistency error, the better view consistency.

\subsection{View Generation Quality}
Table \ref{app-tab:ablation-2d} shows the quantitative results of our text-to-views generation module. As we can see, $prior$-guidance improves the consistency-error from 9.47 to 8.88, and view contrastive learning further improves it to 8.61, indicating both of these two improvements contribute to improving view-consistency. Besides, our complete text-to-views generation module achieves the best consistency-error of 8.56.
Besides, $prior$-guidance also decreases FID and KID, indicating improved view fidelity. This indicates that the extra information $prior$-guidance provided not only improves view-consistency, but also helps the task of view generation. Caption-guidance improves the CLIP-score from 20.62 to 21.22, meanwhile decreasing FID from 15.44 to 14.99. We deem that this improvement stems from the pixel-level supervision signal from CLIP loss compensating for the original token-level training signals. Besides the CLIP loss, the pixel-level $L1$ loss further improves the view fidelity and lowers the FID. We obtain our best text-to-views generation module by integrating all these techniques.

\subsection{3D Object Generation Quality}
Table \ref{tab:app-ablation-2d-3d} reports the CLIP-scores of our views-to-3D generation module for more object categories. As we can see, 3D object generation based on the results of our complete text-to-views generation module achieves the best CLIP-score in all shown categories and achieves the best average CLIP-score among all categories of the ABO test set. Compared with contrastive learning, caption guidance and pixel-level L1 loss, the prior guidance has the largest performance improvement for our 3D-TOGO model. This is because it can greatly improve consistency among the images generated by the text-to-views modules.

\subsection{Qualitative Comparison for $Prior$-guidance and View contrastive learning}

\begin{figure*}[!t]
\centering
\includegraphics[width=0.9\linewidth]{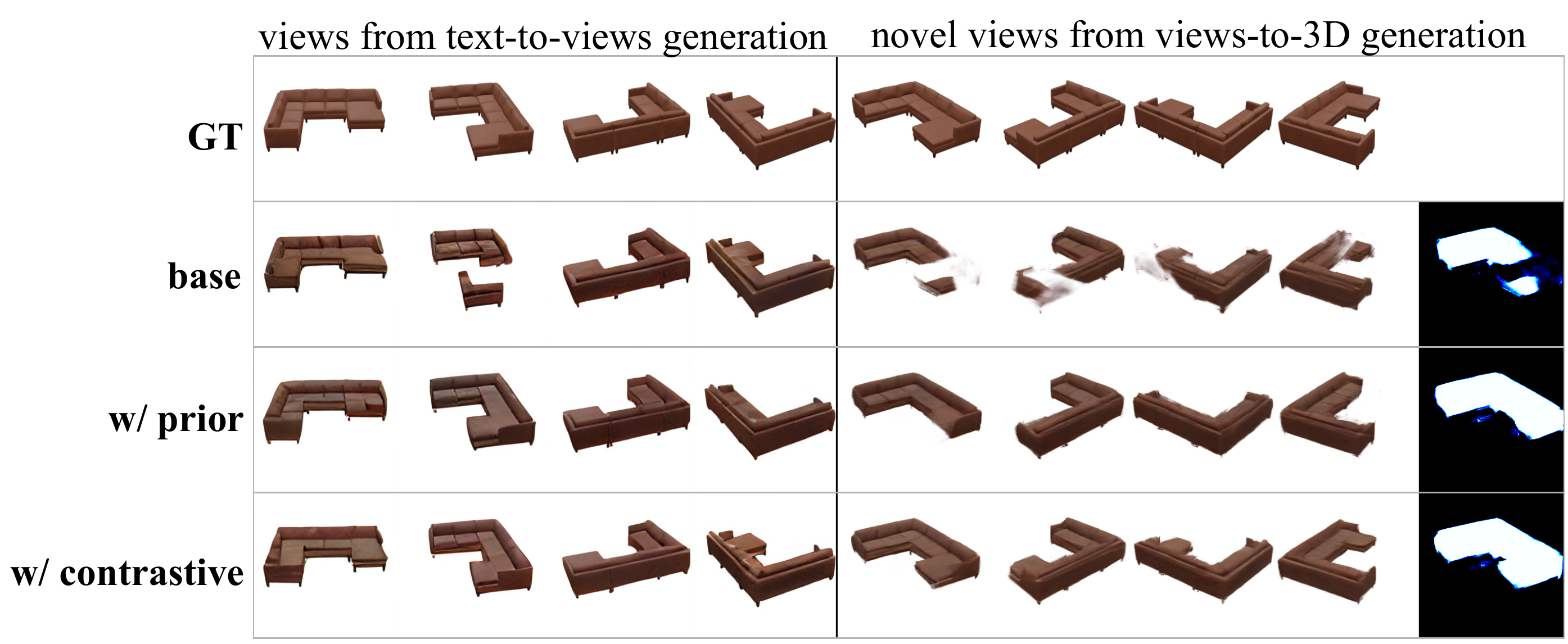}
\vspace{-0.1in}
\caption{Effectiveness of $prior$-guidance and view contrastive learning. 
\textbf{base}, \textbf{prior}, and \textbf{contrastive} indicate base text-to-views generation, $prior$-guidance, and view contrastive learning, respectively. The right-most image is the transmittance from the first generated novel view. 
}
\label{fig:ablation_prior_contrastive}
\end{figure*}

Figure \ref{fig:ablation_prior_contrastive} illustrates the effectiveness of $prior$-guidance and view contrastive learning on improving view consistency. For views from the text-to-views generation module, we can see that compared with the base module, $prior$-guidance improves the view-consistency of 2D generated images, but as the rank of camera pose increases, view-consistency may decrease in more complicated cases due to some accumulated error (see the left 4th image in the 3rd row). Further with the view contrastive learning, the long-range view-consistency is improved. 

\subsection{Qualitative Comparison for Caption Loss}
\begin{figure*}[ht]
\centering
\includegraphics[width=\linewidth]{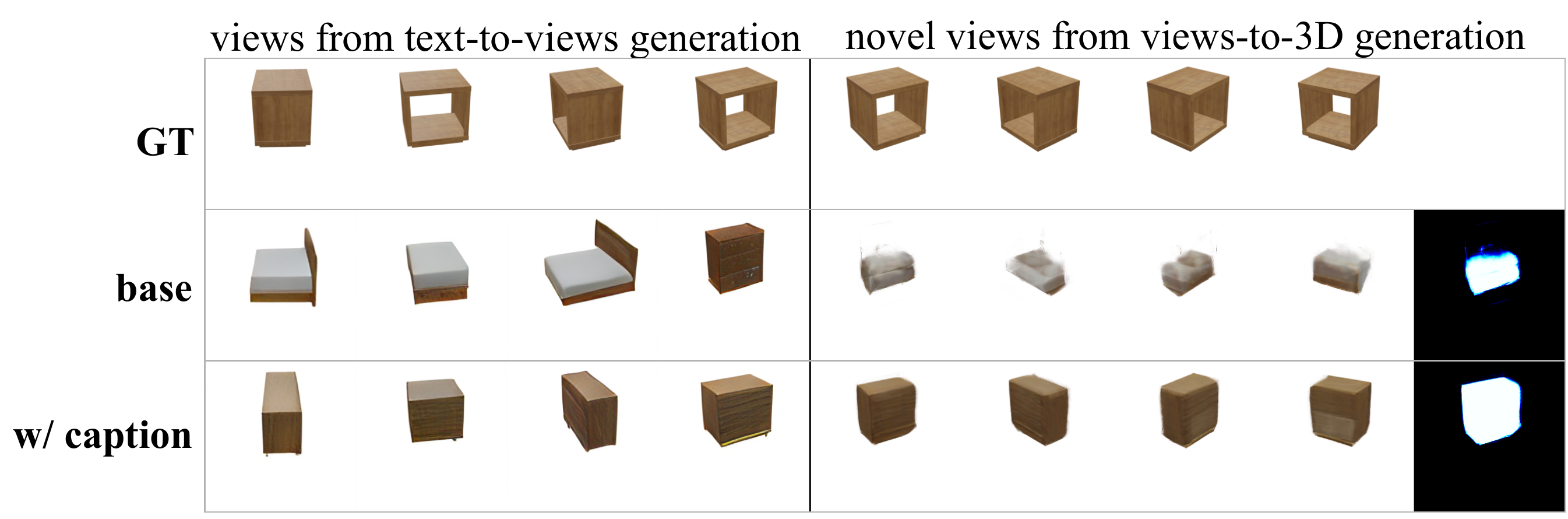}
\caption{Effectiveness of caption loss. 
The input caption is `Oak Finish Industrial Bed side Wood Nightstand'. \textbf{base} and \textbf{caption} indicate base text-to-views generation and caption-guidance, respectively. The right-most image is the transmittance from the first generated novel view. 
}
\label{app-fig:ablation_caption}
\vspace{-0.15in}
\end{figure*}

Figure \ref{app-fig:ablation_caption} shows the comparison between results without caption loss and with caption loss. As we can see, without the caption loss, the model swing between bed and nightstand. On the contrary, with the caption loss, the model consistently generates a nightstand.

\subsection{Qualitative Comparison for L1 Loss}
\begin{figure*}[ht]
\centering
\includegraphics[width=\linewidth]{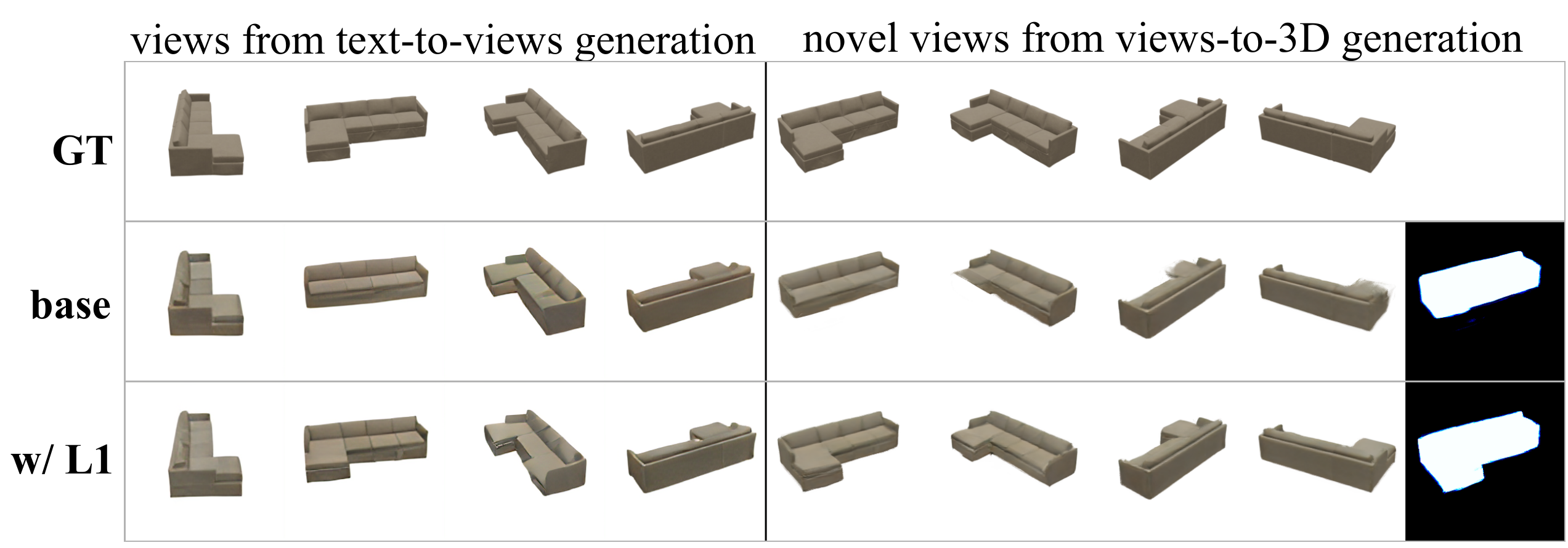}
\caption{Effectiveness of L1 loss. \textbf{base} and \textbf{L1} indicate base text-to-views generation and pixel-level L1 loss, respectively. The right-most image is the transmittance from the first generated novel view.}
\label{app-fig:ablation_l1}
\vspace{-0.15in}
\end{figure*}

Figure \ref{app-fig:ablation_l1} shows the comparison between results without L1 loss and with L1 loss. As we can see, with the L1 loss, a finer-grained training signal, the model can generate results with better details.

\subsection{Multi-views generated by text-to-views module}

\begin{figure*}[ht]
\centering
\includegraphics[width=\linewidth]{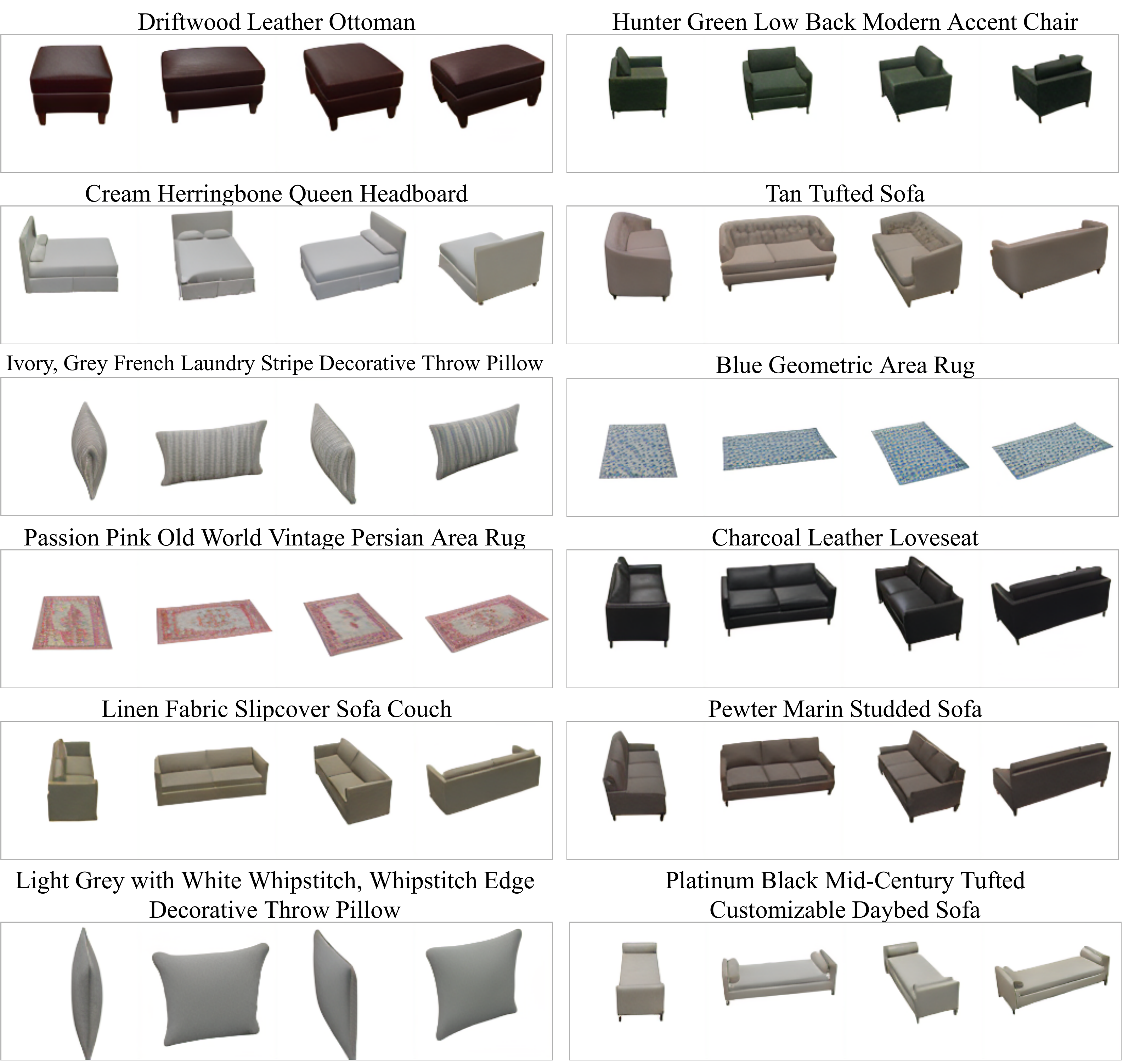}
\caption{Multi-views generated by the text-to-view module for the 12 selected texts.
}
\label{app-fig:2d_results_12texts}
\vspace{-0.15in}
\end{figure*}

It is challenging to generate view-consistent multi-view images from text. Although we improve the consistency among different views by the prior guidance and view contrastive learning, the consistency error of the multi-view images generated by the text-to-views module is still large than that of the ground-truth images as shown in Table 2 of the main text. Figure \ref{app-fig:2d_results_12texts} shows the multi-views generated by the text-to-views module for the 12 selected texts. For those views with high view consistency, a NeRF model can be trained from scratch and regularization is a great idea to improve the quality of the generated 3D objects. However, for those views with small inconsistent contents, such as Headboard and Rug, direct training on these views produces artifacts or even leads to the collapse of the trained model. For the views-to-3D module, we learn scene prior from the ground-truth images and it does not require test-time optimization. Therefore, it can synthesis high-quality 3D objects from the generated views with good efficiency even if there are some small inconsistent contents among the generated views.

\subsection{Generalization of views-to-3D module}
\begin{figure*}[ht]
\centering
\includegraphics[width=\linewidth]{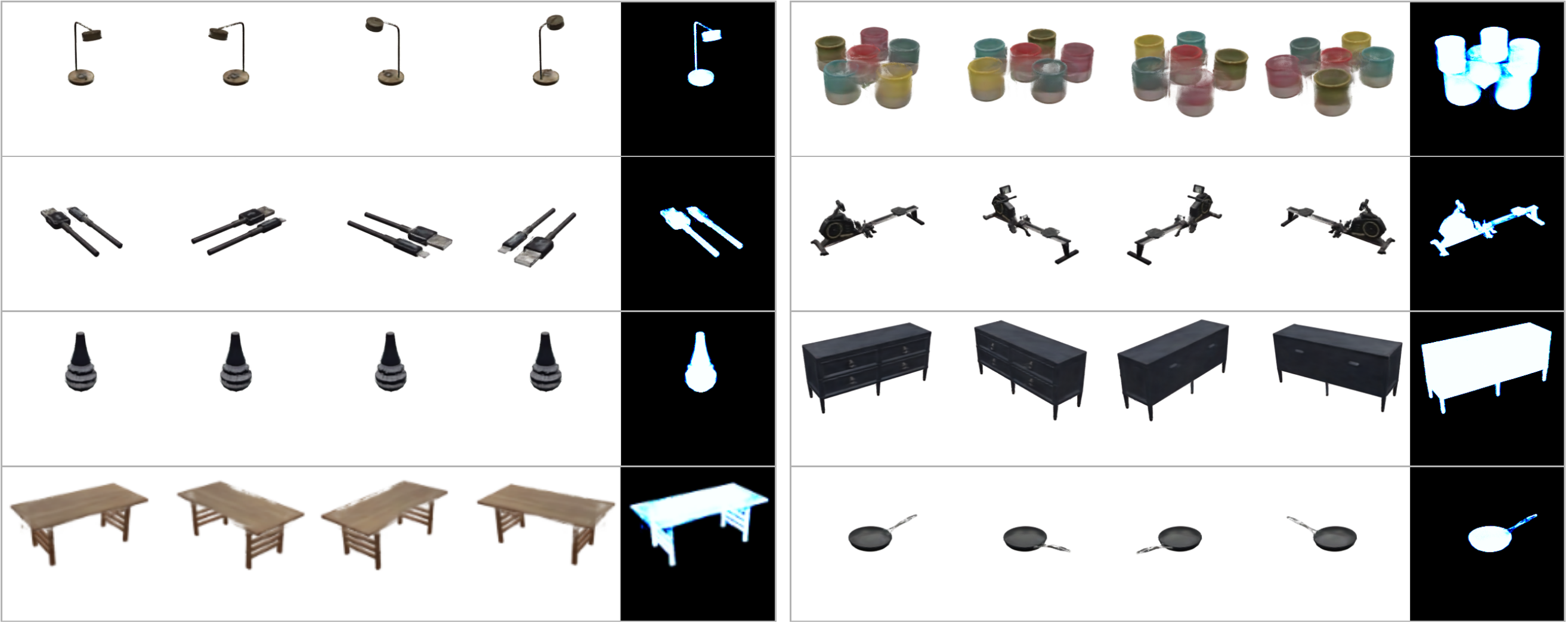}
\caption{Generalization of views-to-3D module.}
\label{app-fig:generalization_of_PixelNeRF}
\vspace{-0.15in}
\end{figure*}

As mentioned above, the views-to-3D module learns a scene prior from the training data, so it has the ability to generalize to unseen categories. Figure \ref{app-fig:generalization_of_PixelNeRF} shows generalization results of the views-to-3D module. Specifically, we train the views-to-3D module on the sofa categories and test the trained model on 8 randomly selected unseen categories. As we can see, the views-to-3D module can obtain good results on the unseen categories.

\subsection{Nearest Neighbor Analysis}
\begin{figure*}[ht]
\centering
\includegraphics[width=\linewidth]{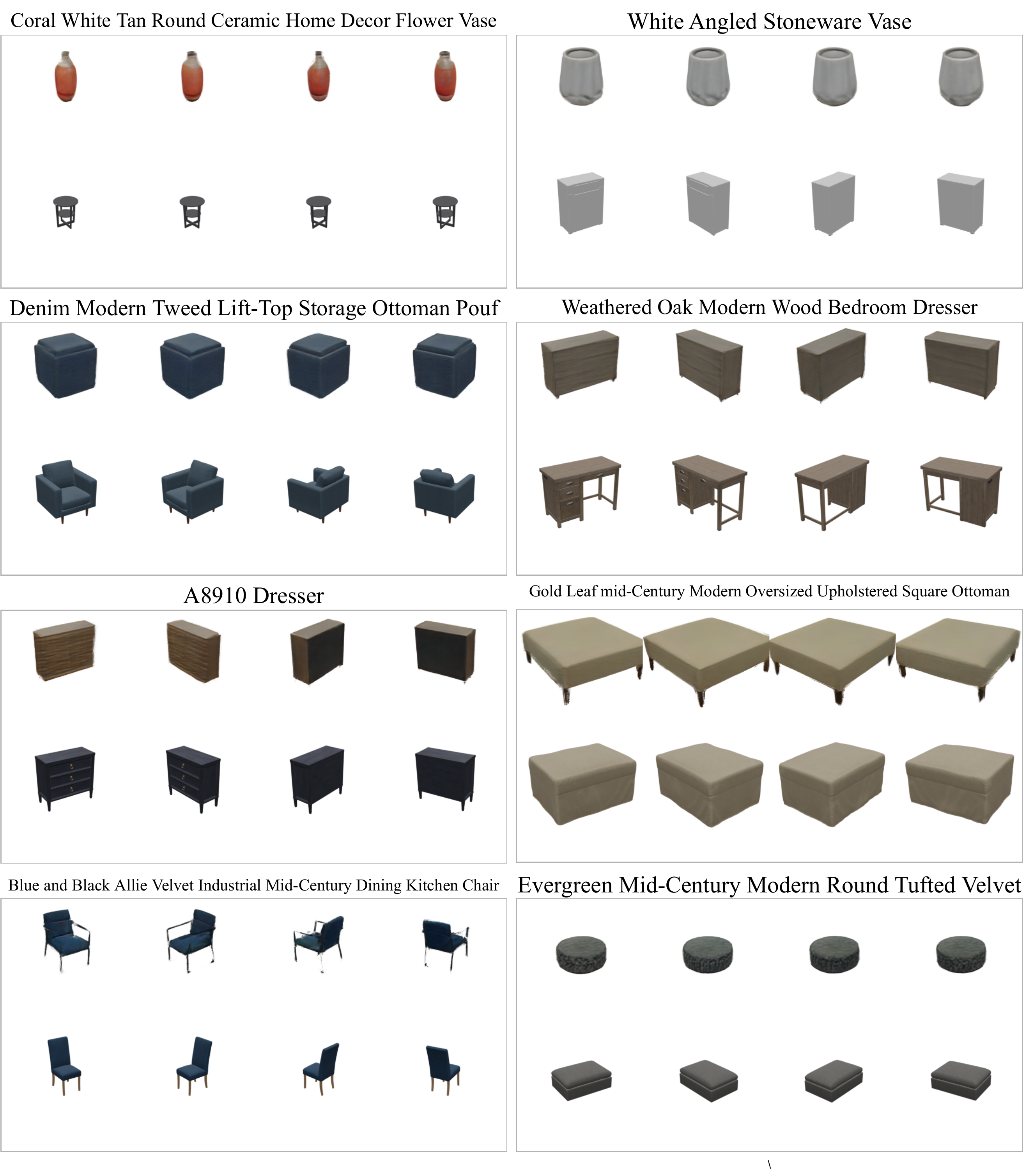}
\caption{Nearest neighbor analysis (1). For each object, the title is the input text, the first row of image are views of the generated object, while the second row of image are views of the closest object in the training set.
}
\label{app-fig:nearest_neighbor_1}
\vspace{-0.15in}
\end{figure*}

\begin{figure*}[ht]
\centering
\includegraphics[width=\linewidth]{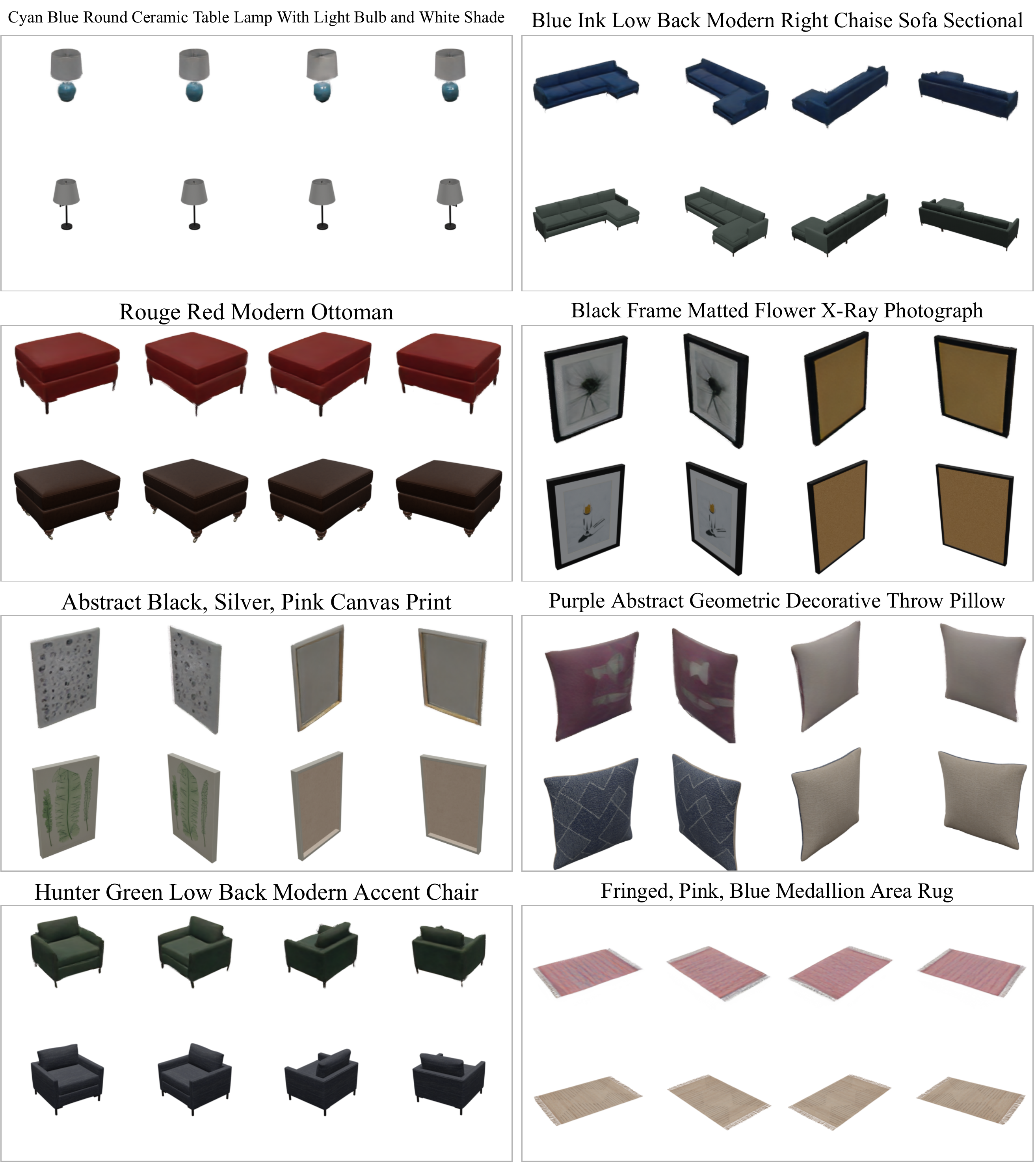}
\caption{Nearest neighbor analysis (2). For each object, the title is the input text, the first row of image are views of the generated object, while the second row of image are views of the closest object in the training set.
}
\label{app-fig:nearest_neighbor_2}
\vspace{-0.15in}
\end{figure*}

We perform nearest neighbor analysis to check whether our model purely memorizes the training data or can generate new objects. Specifically, for each generated object, we calculate the total L2 distance among 4 views between it and each object in the training set and check the closest one. Figure \ref{app-fig:nearest_neighbor_1} and \ref{app-fig:nearest_neighbor_2} shows the results. As we can see, our model does not just memorize the training set, but can generate new objects with different shape and different color according to the input text.

\section{Discussion}
\subsection{Metrics}
Like text-to-image generation, better quantitative metrics are needed for text-guided 3D object generation. NeRF \cite{mildenhall2020nerf} and PixelNeRF \cite{yu2021pixelnerf} use metrics including PSNR, SSIM and LPIPS to measure the average distance between the synthesized novel views and the ground truth views. This is reasonable because their goal is to \textbf{reconstruct} the 3D object from some collected views. However, in this paper, we aim to generate 3D objects from the input caption, without any collected views. In many cases, our model generates 3D objects that are semantically aligned with the input caption but with views different from the `ground truth' views. So, it is unreasonable to use these metrics to measure the performance of our model. We use these metrics in the comparison against text-NeRF following the practice in NeRF. We use CLIP-score in the comparison against baseline methods and the ablation study. CLIP-score measures the semantic similarity between the rendered views from the generated 3D objects and the input caption. Besides, the low image quality also yields a low CLIP-score, so it also reflects the visual fidelity of the generated 3D objects. 
As quantitative metrics are not perfect currently, we also conduct human studies to compare our 3D-TOGO models against baseline methods.

\subsection{3D-TOGO VS  DALLE and CogView}
Using an auto-regressive Transformer for text-to-image generation is commonly used in DALLE\cite{dalle}, Cogview\cite{ding2021cogview} due to its effectiveness in capturing cross-modal correspondence.
Based on this progress, we introduce techniques including Pixel-Level Supervision, Caption-Guidance, prior-Guidance and View Contrastive Learning to better generate consistent multi-view images of an object, which is not considered in previous works and is critical for text-to-3D generation. Besides, DALLE and Cogview focus on the single-view image generation from the input text, while 3D-TOGO focuses on the multi-view images generation from the input text.

\subsection{3D-TOGO VS PixelNeRF}
PixelNeRF\cite{yu2021pixelnerf} is a framework to learn a scene prior for reconstructing NeRF representations from one or a few images. In our method, we introduce PixelNeRF as our views-to-3D module. The reasons we choose PixelNeRF as the backbone of views-to-3D module are three-fold:
\begin{itemize}
    \item PixelNeRF aims to learn a scene prior instead of remembering the training dataset, allowing it to be used for generating objects across different categories or unseen categories. To demonstrate the ability of generalizing to unseen categories, we train the views-to-3D module on the sofa category and test the trained model on 8 randomly selected unseen categories. As shown in Figure \ref{app-fig:generalization_of_PixelNeRF}, the views-to-3D module can obtain good results on the unseen categories.
    \item PixelNeRF is trained across multiple scenes and does not requires test-time optimization, which improves the flexibility for generating 3D objects from text.
    \item PixelNeRF can predict a neural radiance field representation from a spare set of views, which reduces the cost of text-to-views module.
\end{itemize}
PixelNeRF is designed for predicting a Neural Radiance Field representation from few images. In this paper, our task is the cross-category 3D object generation from text. 

\subsection{3D-TOGO VS DreamFields}
Dreamfields\cite{jain2021dreamfields} optimizes a Neural Radiance Field by minimizing the distance between the rendered images and the input text with a pre-trained CLIP model. The pipeline of Dreamfields is completely different from the proposed method. Dreamfields is a one stage solution that goes from text directly to NeRF, while the proposed method is a two stage solution. In the proposed method, we first generate consistent multi-view images of an object and then reconstruct 3D object from these generated consistent multi-view images. The rendered images of Dreamfields are rendered by sampling different camera poses, while the multi-view images of the proposed method are directly generated from the input text. Inherited from NeRF\cite{mildenhall2020nerf}, Dreamfields requires training a separate model for each input text, which usually takes more than 1 hour with 8 TPU cores. Compared with Dreamfields, the proposed method has an efficient generation process. Besides, hyperparameters of Dreamfields are manually tuned on the COCO dataset. Given a caption of the ABO dataset, it is difficult to do such hyperparameter tuning. Another advantage of 3D-TOGO is that it can generate more realistic 3D objects.

\subsection{3D-TOGO VS CLIP-Mesh and DreamFusion}

CLIP-Mesh \cite{khalid2022clip} and DreamFusion \cite{poole2022dreamfusion} seem to produce amazing and promising 3D objects. However, both of them require training a separate model for each input text on the fly, which is time-consuming and limits its practical application. Compared with them, 3D-TOGO is a generic model and does not require inference time optimization. 3D-TOGO can generate a 3D object within 2 minutes, while CLIP-Mesh and DreamFusion require 50 minutes and 3 hours respectively. 

\section{Failure cases}
\begin{figure*}[ht]
\centering
\includegraphics[width=\linewidth]{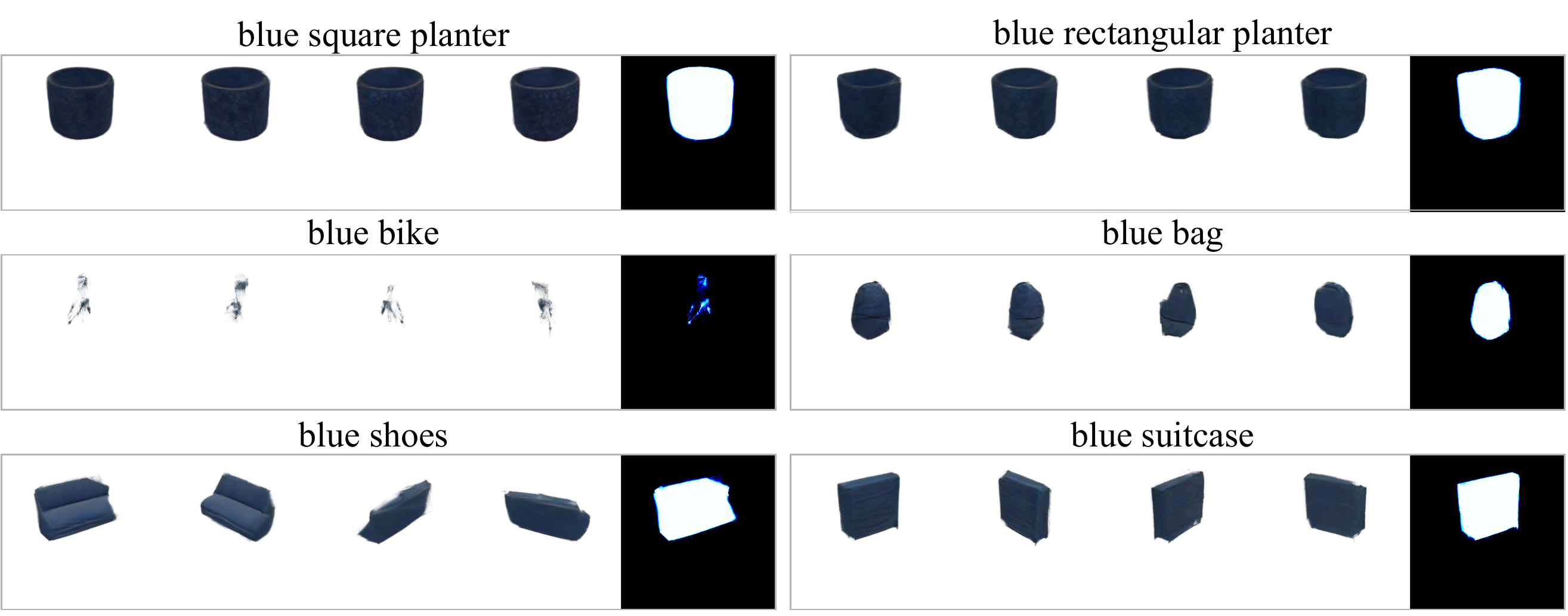}
\caption{Failure cases.}
\label{app-fig:failure_cases}
\vspace{-0.15in}
\end{figure*}

\begin{figure*}[ht]
\centering
\includegraphics[width=1.0\linewidth]{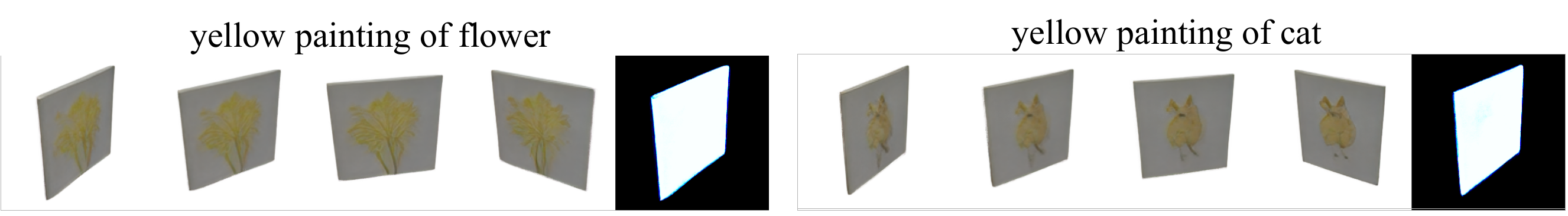}
\caption{Flower and cat.}
\label{app-fig:flower_and_cat}
\vspace{-0.15in}
\end{figure*}

Figure \ref{app-fig:failure_cases} shows some failure cases of our model. If the input text is completely out of the distribution of ABO, it might fail to generate desired text-matched 3D objects. 
In the experiments, one interesting thing is that our 3D-TOGO can generate 3D objects of unseen categories by using prompt engineering "a painting of" as shown in Figure \ref{app-fig:flower_and_cat}.
We believe the generalization ability of our 3D-TOGO model can be scaled by pretraining on more text-image samples and using a larger generation model.  CO3D\cite{reizenstein2021common}, containing a total of 1.5 million frames from nearly 19,000 videos capturing objects from 50 MS-COCO categories, is one that could be used to improve the ability of our 3D-TOGO model. Besides, it is possible to improve the generalization of the proposed method with single-view images of objects, which are easily collected.
In some cases, 3D-TOGO could be unable to control the shape of generated 3D objects, which can be solved by using repetition words describing 3D shapes.

\section{Conclusion}
\subsection{Potential negative social impact}
Our method has no ethical risk on dataset usage and privacy violation since all the benchmarks are publicly available. Besides, our method offers a new way to generate 3D objects and we do not expect a negative social impact as the generated 3D objects are generated with textual guidance. If the training data on some sensitive categories is available, the proposed method could be misused for generating real 3D humans or weapons. Such misuse of 3D object generation from text techniques poses a societal threat, and we do not condone using our work with the intent of spreading misinformation or tarnishing reputation.

\subsection{Limitations and future works}

If the input text is completely out of the distribution of ABO, it might fail to generate desired text-matched 3D objects. In the further, we plan to improve the generalization of the proposed method with single view images of objects, which are easily collected. Besides, we will apply our 3D-TOGO model to datasets from other domains, such as humans.

\end{appendices}

\end{document}